%% file: main.tex
\documentclass[twoside]{article}

\usepackage[preprint]{aistats2027}

\usepackage[round]{natbib}

\usepackage{graphicx}
\usepackage[utf8]{inputenc} 
\usepackage[T1]{fontenc}    
\usepackage{float}          
\usepackage{hyperref}       
\usepackage{url}            
\usepackage{booktabs}       
\usepackage{amsfonts}       
\usepackage{nicefrac}       
\usepackage{microtype}      
\usepackage{xcolor}         

\input{preamble}

\begin{document}

\twocolumn[

\aistatstitle{Stable and Calibrated Latent Force ODE Filters}

\aistatsauthor{ Paul Fischer \And Jonathan Schmidt \And Philipp Hennig }

\aistatsaddress{ T\"ubingen AI Center \\ University of T\"ubingen \And T\"ubingen AI Center \\ University of T\"ubingen \And T\"ubingen AI Center \\ University of T\"ubingen } ]

\begin{abstract}
Estimating unobserved forces in mechanistic systems is a central task of scientific machine learning.
Approximate Gaussian filtering methods that synthesize mechanistic and empirical information have shown remarkable capacity in jointly identifying latent processes together with the system state in a single forward-backward pass.
The resulting latent force ODE filter framework channels information from observational data specifically into the unknown parts of the system.
This work addresses open questions about calibration and stability of the method.
We show that the joint posterior over state and force is the reweighted conventional latent force posterior.
This uncovers a crucial dependence of the stability of the method on the prior variance of the system state, which effectively controls the flow of information from observations to the latent force through the mechanistic system.
From this insight, we construct an improved algorithm that provides stable and calibrated plug-and-play inference.
We demonstrate the improved framework on several simulated experimental settings.
\end{abstract}

\section{Introduction}

\begin{figure*}[t]
  \centering
  \includegraphics[width=\textwidth]{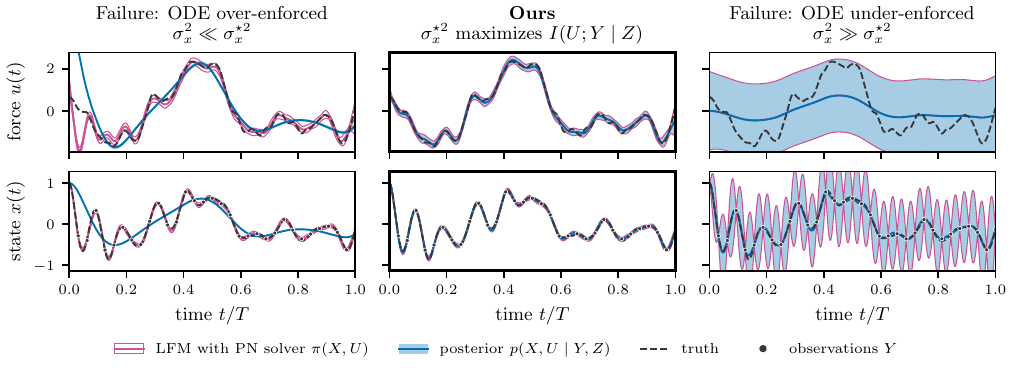}
  \caption{\textbf{The state diffusion controls the latent force inference.} Latent force ODE filter on the Duffing oscillator of Eq.~\eqref{eq:duffing} with the position observed (Appendix~\ref{app:state-diffusion}). Center: the diffusion $\sigma_x^{\star 2}$ that maximizes $I(U;Y\mid Z)$ (Sec.~\ref{subsec:info-criterion}). Left and right: $10^{8}$ times smaller and $10^{6}$ times larger. Bands: mean $\pm 2$ standard deviations of the classical latent force model with the probabilistic numerical solver, $\pi(X,U)$, and of the full latent force ODE filter posterior $p(X,U\mid Y,Z)$.}
  \label{fig:failure-directions}
\end{figure*}

Many dynamical systems are driven by external forces that cannot be observed directly.
Examples include wind loads acting on a bridge, stimulus currents driving a neuron, and contact forces disturbing a robot.
Downstream decision making requires uncertainty quantification and identification of such unknown inputs.
Hybrid models, which combine mechanistic knowledge in the form of differential equations with function approximations learned from data, leverage the advantages of both: the expressivity of function approximation and the generalizability of mechanistic models.

Latent force models (LFMs) \citep{alvarez2009latent,hartikainen2012state} are hybrid models that place a Gaussian process (GP) prior on the force, integrate it through the differential equation into a joint input-state prior, and condition this prior on the data.
For systems described by ordinary differential equations (ODEs), \citet{schmidt2021probabilistic} introduced an adjacent approach, which we refer to as latent force ODE filters.
It builds on ODE filters \citep{tronarp2019probabilistic,kersting2020convergence}, probabilistic numerical solvers that treat the local gradient information of the vector field as data within a Bayesian inference framework \citep{cockayne2019bayesian,hennig2022probabilistic}.
State and force are modeled as independent GPs, and the mechanistic information only enters at inference time, essentially treating the dynamics model as data.
Adding empirical observations then lifts the solver to a hybrid model.
In ODE filters, the mechanistic information is encoded as \emph{pseudo-data} pinned to zero, not a real measurement of the system.
Conditioning confines the prior to the restricted measure that is consistent with the dynamics \citep{da2025constructive}.
Pinning the residual to zero is harmless while it only serves to solve the dynamics, but consequential once it is used as evidence for calibration and model selection.
For latent force ODE filters, these aspects have not been studied yet.
Model selection and calibration thus remain open challenges, hindering real-world applications and comparison to competing hybrid methods.

\paragraph{Contributions}
This paper turns the pseudo-data distinction into a calibration principle for latent force ODE filters.
We show that the state diffusion of the prior process not only scales the numerical uncertainty but also controls the rate at which the latent force learns from the residual.
This dual role leads to two failure modes (Fig.~\ref{fig:failure-directions}).
A small diffusion enforces the dynamics too rigidly: the force compensates for the limited expressivity of the state prior, uncertainty bands contract, and the posterior fails to model the data.
A large diffusion decouples force and state: the force is poorly inferred and the state interpolation is massively underconfident.
Selecting the diffusion from the residual evidence, as is common for ODE filters, over-enforces the dynamics.
Selecting it instead by the information the data carry about the force, given the ODE, avoids both failures and yields calibrated and accurate posteriors.

Our main contributions are:
\begin{itemize}
  \item We show that the posterior of a latent force ODE filter factorizes into a classical latent force model posterior and a Gibbs weight on the force whose temperature is the state diffusion. Latent force inference fails for both too high and too low values of the prior diffusion (Secs.~\ref{subsec:posterior-factorization}--\ref{subsec:failure-directions}).
  \item We propose selecting the diffusion by the mutual information $I(U;Y\mid Z)$ of force and data given the ODE, and evaluate it within an algorithm that refines the grid and calibrates hyperparameters (Secs.~\ref{subsec:info-criterion} and~\ref{subsec:calibration-guide}).
  \item On three nonlinear systems, selecting the diffusion by $I(U;Y\mid Z)$ recovers the force on every problem from a single start (Sec.~\ref{sec:experiments}).
\end{itemize}

\section{Background}

\subsection{ODE filters}\label{subsec:ode-filters}

ODE filters are probabilistic solvers of initial value problems
\begin{equation}
  x^{(m)} = f\big(x(t), \dot x(t), \dots, x^{(m-1)}(t), t\big),
  \label{eq:ivp}
\end{equation}
with $x^{(m)} = \mathrm d^m x(t)/\mathrm d t^m$, known initial values $x(0), \dots, x^{(m-1)}(0)\in\Rd$ and $f: \R^{dm}\times [0, T] \to \Rd$.
For notational simplicity we write the scalar first-order case, $d=m=1$.
ODE filters treat numerical integration as sequential Bayesian inference \citep{tronarp2019probabilistic} and compute a posterior over solutions that quantifies the numerical error due to limited compute,
\begin{equation}\label{eq:ODE-IVP_posterior}
 p\left(x(t)\mid x(0)=x_0, \left\{\dot{x}(t_n) = f(x(t_n), t_n)\right\}_{n=1}^N\right).
\end{equation}

\paragraph{Prior.}
Smoothness assumptions are encoded by a Gauss--Markov prior on the extended state $X(t) = [x(t), \dot{x}(t), \ldots, x^{(q)}(t)]^\top \in \R^{d(q+1)}$, usually the integrated Wiener process IWP($q$) that models $x^{(q)}$ as Brownian motion.
Its Markov property gives the discrete state space form
\begin{equation}
  p(X_{n+1} \mid X_n) = \N(A_h X_n, Q_h),
\end{equation}
with $A_h$ and $Q_h = \sigma^2\check{Q}_h$ in closed form for the order $q$, step size $h = t_{n+1}-t_n$ and diffusion $\sigma^2$ \citep{schober2019probabilistic}, which enables linear-in-time filtering and smoothing \citep{hartikainen2010kalman}.
The order $q$ sets the convergence rate \citep{kersting2020convergence}.

\paragraph{Data.}
The ODE enters as pseudo-data $\{z_n = 0 \}_{n=1}^N$ on the residual at the grid nodes $t_1 < \dots < t_N$,
\begin{equation}
  \begin{aligned}
    \mathcal{Z}[X](t) &= \dot{x}(t) - f(x(t), t),\\
    p(z_n \mid X_n) &= \N(\mathcal{Z}[X](t_n), \Rode),
  \end{aligned}
\end{equation}
where $\Rode$ is typically zero but can represent model misspecification or linearization error \citep{kersting2016active,kersting2020convergence}.
Nonlinear $f$ are handled by approximate filters \citep{sarkka2023bayesian}.

\paragraph{Calibration.}\label{sec:ode-filter-calibration}
With $\Rode=0$, a time-invariant diffusion $\sigma^2$ scales the whole posterior covariance and is calibrated by its quasi-maximum likelihood estimate under the residual evidence $p(z_{1:N}) = \prod_{n=1}^N p(z_n \mid z_{1:n-1}) = \prod_{n=1}^N \N(z_n \mid \hat z_n, S_n)$.
A time-varying diffusion allows online calibration and adaptive step-size selection \citep{bosch2021calibrated}.

\subsection{Latent force models}

Consider a dynamical system driven by an unknown input, disturbance, or latent force $u(t)$, observed on a discrete time grid through a measurement model $h$,
\begin{equation}\label{eq:lfm-problem}
  \begin{split}
    \dot{x}(t) &= f(x(t), u(t), t), \\
    y_k &= h(x(t_k),t_k) + \epsilon_k,
  \end{split}
\end{equation}
where measurements are corrupted by additive Gaussian noise $\epsilon_k \sim \N(0, \Robs)$.
The latent force inference problem considered hereafter is to identify an open-loop input $u^\star(t)$ that generates a trajectory $x^\star (t)$ observed through the measurements $\{y_k\}_{k=1}^K$, formulated as Bayesian inference on the joint input-state posterior $p(x(t), u(t) \mid \{y_k\}_{k=1}^K, f)$.
Latent force models \citep{alvarez2009latent,hartikainen2012state} construct a joint input-state prior by integrating the input prior through the ODE $p(x, u) = p(x \mid u)p(u)$.

\subsection{Latent force ODE filters}\label{sec:lf-ode-filters}
\begin{figure}[ht]
  \centering
  \begin{tikzpicture}[
    >={Latex[length=2.2mm]},
    node/.style={circle, draw, thick, fill=white, minimum size=0.95cm, inner sep=1pt},
    obs/.style={circle, draw, thick, fill=black!15, minimum size=0.95cm, inner sep=1pt},
    arr/.style={->, thick},
    coupling/.style={->, very thick, densely dashed, darkblue},
    group/.style={draw, rounded corners, dashed, inner sep=4mm},
    grouplabel/.style={font=\small\itshape, text=black!60},
    platecount/.style={font=\small, text=black!60, inner sep=1.5pt, anchor=south east},
  ]
    \node[node] (tx) at (0, 2.4) {$\varphi_x$};
    \node[node] (X)  at (3.0, 2.4) {$X$};
    \node[obs]  (Y)  at (6.0, 2.4) {$Y$};
    \node[node] (tu) at (0, 0.6) {$\varphi_u$};
    \node[node] (U)  at (3.0, 0.6) {$U$};
    \node[obs]  (Z)  at (6.0, 0.6) {$Z$};

    \begin{scope}[on background layer]
      \node[group, fit=(tx)(tu), label={[grouplabel]above:{$\varphi$}}] {};
      \node[group, fit=(X)(U), label={[grouplabel]above:{$\tilde{X}$}}] (plateX) {};
      \node[group, fill=black!5, fit=(Y)(Z), label={[grouplabel]above:{$\tilde{Y}$}}] (plateY) {};
      \node[group, inner sep=2.5mm, fit=(Y)] (plateK) {};

      \draw[arr] (tx) -- (X);
      \draw[arr] (X)  -- (Y);
      \draw[arr] (X)  -- (Z);
      \draw[arr] (tu) -- (U);
      \draw[arr] (U)  -- (Z);
      \draw[coupling] (U) -- (X) node[midway, left] {$f$};
    \end{scope}

    \node[platecount] at ([shift={(-1.5mm,1.5mm)}]plateX.south east) {$N$};
    \node[platecount] at ([shift={(-1.5mm,1.5mm)}]plateY.south east) {$N$};
    \node[platecount] at ([shift={(-1.0mm,1.0mm)}]plateK.south east) {$K$};
  \end{tikzpicture}
  \caption{\textbf{Directed graphical model of the latent force ODE filter.} State and force are coupled only through the residual pseudo-observation and the initial state. $\varphi_x$, $\varphi_u$: prior hyperparameters. Shaded: observed. Plates: $N$ grid nodes and $K\le N$ observation times. \couplingarrow: mechanism $f$, replaced at inference by $Z$.}
  \label{fig:lf-ode-filter-graphical-model}
\end{figure}
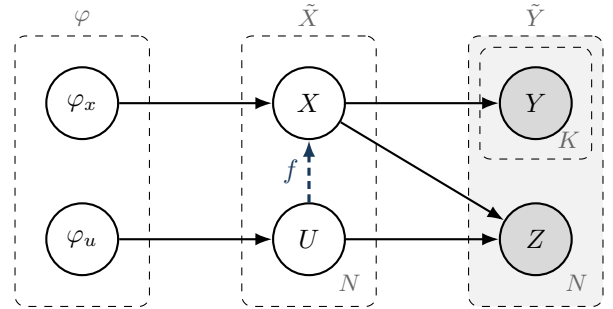

Latent force ODE filters \citep{schmidt2021probabilistic}, in turn, place independent Gauss--Markov priors on extended inputs and states, with state space form:
\begin{equation}
  \begin{split}
    p(X_{n+1} \mid X_n) &= \N(A_h X_n, Q_h),\\
    p(U_{n+1} \mid U_n) &= \N(\bar{A}_h U_n, \bar{Q}_h).
  \end{split}
\end{equation}
The extended states $X$ and $U$ are obtained by appending $x$ and $u$ with their first $q_x$ and $q_u$ derivatives, respectively.
We write $\sigma_x^2$ for the diffusion of the state prior, $Q_h=\sigma_x^2\check Q_h$, and $\varphi=(\varphi_x,\varphi_u)$ for the hyperparameters of state and force prior, with $\varphi_x=\sigma_x^2$.
Any Gauss--Markov prior is admissible.
The integrated Wiener process is the default state prior, as for ODE filters (Sec.~\ref{subsec:ode-filters}), while the force prior can benefit from additional structure that makes the inverse problem more identifiable \citep{stuart2010inverse}.
Inference is performed by conditioning on the sparse noisy observations $\{y_k\}_{k=1}^K$ and the dense ODE residuals $\{z_n = 0\}_{n=1}^N$, with observation models:
\begin{equation}\label{eq:lf-ode-filter-ssm}
  \begin{split}
    p(z_n \mid X_n, U_n) &= \N\big(\mathcal{Z}[X,U](t_n), \Rode\big),\\
    p(y_k \mid X(t_k)) &= \N\big(h[X](t_k), \Robs\big).
  \end{split}
\end{equation}
The residual grid is finer than the measurement grid, $N \ge K$, and contains the measurement times, $\{t_k\}_{k=1}^K \subseteq \{t_n\}_{n=1}^N$.
The specific $N$ is a modeling choice and can in principle be selected adaptively to control the numerical error \citep{bosch2021calibrated}.
For notational simplicity we abbreviate $Y \equiv \{y_k\}_{k=1}^K$ and $Z \equiv \{z_n\}_{n=1}^N$, as well as $X \equiv X_{0:N} \equiv [X_0, \cdots, X_N]$, and $U \equiv U_{0:N}$ accordingly.
\citet{schmidt2021probabilistic} summarize state and force into an extended state $\tilde{X}=(X,U)$ and model and observations into generalized observations $\tilde{Y} = (Y,Z)$ (Fig.~\ref{fig:lf-ode-filter-graphical-model}), and compute the joint posterior $p\big(\tilde{X} \mid \tilde{Y}\big)$ in a single extended Kalman smoother pass.
Performing the observation update after the ODE update instead splits the joint log-likelihood into a residual and an observation term:
\begin{equation}\label{eq:prediction-error-decomposition}
  \begin{split}
    \log p(Y,Z \mid \varphi) = &\underbrace{\sum_{n=1}^N \log p(z_n \mid z_{1:n-1}, y_{1:k(n-1)}, \varphi)}_{\lode}\\
    +
    &\underbrace{\sum_{k=1}^K \log p(y_k \mid y_{1:k-1}, z_{1:n(k)}, \varphi)}_{\lobs}.
  \end{split}
\end{equation}
Here, $k(n)$ is the number of observations up to $t_n$, and $n(k)$ the grid index of $t_k$.
The two sums are log-likelihoods, the residual evidence $\lode$ and the observation evidence $\lobs$.
The remainder of the paper assumes the following setting:
State priors are IWPs and force priors Matérn processes (Appendix \ref{bsec:matern}).
All model misspecification is absorbed into the latent force $\Rode=0$.
The observation noise $\Robs$ and force smoothness $q_u$ are known.
Calibration thus requires selecting state smoothness $q_x$, state diffusion $\sigma_x^2$, force length scale and variance $\varphi_u=(\ell_u,\sigma_u^2)$ and the grid density $\kappa = N/K$.
For the full predict and update equations see Appendix \ref{app:update-equations}.

\section{Method}

The latent force ODE filter posterior factorizes into a LFM posterior and a weight that measures how much the residual updates the force (Sec.~\ref{subsec:posterior-factorization}).
The weight arises from the ODE filter construction, which conditions on the mechanistic information.
The state diffusion enters both factors (Sec.~\ref{subsec:state-diffusion}): it sets the numerical uncertainty in the LFM factor and is the temperature of the weight.
Latent force inference fails at both ends (Sec.~\ref{subsec:failure-directions}, Fig.~\ref{fig:failure-directions}).
For a small diffusion, the weight selects the force from the residual alone and the data are ignored.
For a large diffusion, the LFM factor no longer depends on the force and the data cannot inform it.
In Section~\ref{subsec:info-criterion} we argue that both failures occur when the mutual information between the data and the force under the model decreases.
As a consequence, the state diffusion loses its interpretation from classical ODE filters, and calibration results do not transfer.
We therefore propose an algorithm for latent force ODE filters in Section~\ref{subsec:calibration-guide} that serves as scaffolding to study different calibration objectives (Sec.~\ref{subsec:info-criterion}).

\subsection{Posterior factorization}\label{subsec:posterior-factorization}

The full posterior of the graphical model in Fig.~\ref{fig:lf-ode-filter-graphical-model} factorizes into a classical LFM posterior $\pi$ reweighted by a factor $\mathcal W(U)$: $p(X,U\mid Y,Z)=\pi(X,U)\mathcal W(U)$.
Conditioning on the residual $Z$ first gives
\begin{equation}\label{eq:one-factor}
  \begin{aligned}
    p&(X,U \mid Y,Z)
      = \frac{p(Y \mid X)\, p(Z \mid X,U)\, p(X,U)}{p(Y,Z)}\\
      &= \underbrace{\frac{p(Y \mid X)\, p(X \mid U,Z)\, p(U)}{\tilde p(Y)}}_{\pi(X,U)}\;
         \underbrace{\frac{\tilde p(Y)}{p(Y \mid Z)}\,\frac{p(U \mid Z)}{p(U)}}_{\mathcal{W}(U)},
  \end{aligned}
\end{equation}
with
\begin{equation*}
  \begin{aligned}
    &\tilde p(Y) = \int p(Y \mid U,Z)\,p(U)\,\mathrm{d}U,\\
    &\mathcal W(U) = \frac{\mathcal V(U;Z)}{\mathbb E_{\pi}\left[\mathcal V(U;Z)\right]},
    \quad
    \mathcal V(U;Z) = \frac{p(U \mid Z)}{p(U)}.
  \end{aligned}
\end{equation*}

Conditionals on $Z$ can be read as ``given the ODE''.
$\pi(X,U)$ is the posterior of a latent force model \citep{hartikainen2012state}, in which the deterministic ODE solution $U \mapsto X$ is replaced by the probabilistic numerical solution $p(X \mid U, Z)$.
The weight $\mathcal W$ measures how much the force prior changes through the ODE observations $Z$.

\subsection{State diffusion}\label{subsec:state-diffusion}

The state diffusion $\sigma_x^2$ enters both factors of Eq.~\eqref{eq:one-factor}: it scales the numerical uncertainty in $\pi(X,U)$, and it is the temperature of $\mathcal W(U)$.
We show this for a single residual update from an independent prior on state and force, with the residual linearized as $\mathcal Z[X,U]=CX+BU+\hat z$ (derivation in Appendix~\ref{app:single-update}).
The predicted state covariance $\pre{P_{xx}}$ and the state part of the innovation covariance, $S_x=C\pre{P_{xx}}C^\top$, are both proportional to $\sigma_x^2$.
Given the force, the numerical solution inside $\pi(X,U)$ is
\begin{equation}\label{eq:numerical-solution}
  p(X\mid U,z=0) = \N\big(X\mid -L_{x\mid u}(BU+\hat z),\ \ode{P_{x\mid u}}\big),
\end{equation}
with gain $L_{x\mid u}=\pre{P_{xx}}C^\top S_x^{-1}$.
The diffusion cancels in the gain, so, as in a classical ODE filter, it leaves the mean of the solution unchanged and scales its covariance $\ode{P_{x\mid u}}$.
The weight depends on the force through the likelihood of the residual with the state marginalized, $p(z=0\mid U)=\N(0\mid BU+\hat z,\, \sigma_x^2 \check S_x)$:
\begin{equation}\label{eq:gibbs-weight}
  \begin{aligned}
    \mathcal W(U) &\propto \exp\!\left(-\frac{\mathcal E(U)}{2\sigma_x^2}\right),\\
    \mathcal E(U) &= (BU+\hat z)^\top \check S_x^{-1}(BU+\hat z).
  \end{aligned}
\end{equation}
The weight therefore takes the form of a Gibbs distribution with temperature $\sigma_x^2$ and energy $\mathcal E(U)$.

\subsection{Failure directions}\label{subsec:failure-directions}

The two roles of the diffusion conflict, and in both limits the data stop informing the force, each time through one of the two factors (Fig.~\ref{fig:failure-directions}).
\paragraph{Hard enforcement: $\sigma_x^2\to0$.}
The numerical uncertainty vanishes, and $\pi(X,U)$ tends to a classical latent force model with the solution map $U\mapsto-L_{x\mid u}(BU+\hat z)$, which can fit the data.
The weight $\mathcal{W}$, however, concentrates on the forces of least energy $\mathcal E(U)$, which the residual alone determines.
The posterior of the force therefore no longer depends on the data, although $\pi$ would fit them.
In Fig.~\ref{fig:failure-directions} (left), $\pi$ follows the truth, while the posterior departs from the data over the first part of the window with narrow bands.
\paragraph{Weak enforcement: $\sigma_x^2\to\infty$.}
The weight $\mathcal{W}$ tends to one, and the posterior to $\pi(X,U)$, which itself depends on the diffusion: the covariance of the numerical solution grows with $\sigma_x^2$, while its dependence on the force stays fixed.
The likelihood $p(Y\mid U,Z)$ of the force in $\pi$ therefore becomes flat.
The state explains the data on its own, and the force keeps its prior.
In Fig.~\ref{fig:failure-directions} (right), the posterior coincides with $\pi$, the force keeps its prior width, and the state follows the observations with bands that widen between them.

In the filter, the gain with which an observation updates the force vanishes in proportion to $\sigma_x^2$ in the first limit and to $\sigma_x^{-2}$ in the second (Appendix~\ref{app:single-update}, Table~\ref{tab:diffusion-limits}).

\subsection{Selecting the diffusion}\label{subsec:info-criterion}

Table~\ref{tab:criteria} lists the candidate objectives considered for selecting the state diffusion $\sigma_x^2$.
They form two pairs, one scoring the evidence of the data and one the information the data carry about the force.

\begin{table}[ht]
  \caption{Criteria $\mathcal J$ for the state diffusion (line~\ref{alg:fit} of Algorithm~\ref{alg:calibration-guide}): evidence (top) and information (bottom), causal member first, with the cost of one evaluation in filter passes.}
  \label{tab:criteria}
  \centering
  \small
  \vspace{2pt}
  \begin{tabular}{@{}lll@{}}
    \toprule
    Criterion $\mathcal J$ & Definition & Passes \\
    \midrule
    $\lobs$ & Eq.~\eqref{eq:prediction-error-decomposition} & 1 \\
    $\log p(Y\mid Z)$ & $\lode+\lobs-\log p(Z)$ & 2 \\
    \midrule
    $D_\xi$ & Eq.~\eqref{eq:directed-information} & 2 \\
    $I_\xi(U;Y\mid Z)$ & Eq.~\eqref{eq:mutual-information-gaussian} & 4 \\
    \bottomrule
  \end{tabular}
\end{table}

\paragraph{Evidence.}
The observation evidence $\lobs$ of Eq.~\eqref{eq:prediction-error-decomposition} is the sum of the filter's one-step-ahead predictive log-likelihoods of the observations.
The conditional evidence $\log p(Y\mid Z)=\log p(Y,Z)-\log p(Z)$ conditions every observation on all residuals and has been used for parameter inference \citep{tronarp2022fenrir,wu2024data}.
Both score how well the model fits the data.

\paragraph{Information.}
In both failure directions the data stop informing the force (Sec.~\ref{subsec:failure-directions}).
How much they inform it given the ODE is measured by the mutual information of $U$ and $Y$ under $p(U,Y\mid Z=0)$ \citep{cover2006elements},
\begin{equation}\label{eq:mutual-information}
  \begin{aligned}
    I(U;Y\mid Z) &= \mathbb E\!\left[\log\frac{p(U\mid Y,Z=0)}{p(U\mid Z=0)}\right]\\
    &= H(U\mid Z=0)-H(U\mid Y,Z=0),
  \end{aligned}
\end{equation}
with the expectation under $p(U,Y\mid Z=0)$ and $H$ the (conditional) entropy.
The mutual information is the expected reduction in the uncertainty about the force when the data are observed in addition to the ODE.
For the model linearized at the trajectory $\xi$, all these conditionals are Gaussian with covariances that do not depend on the values conditioned on, and
\begin{equation}\label{eq:mutual-information-gaussian}
  I_\xi(U;Y\mid Z)=\frac12\log\frac{|\operatorname{Cov}_\xi(Y\mid Z=0)|}{|\operatorname{Cov}_\xi(Y\mid Z=0,U)|}.
\end{equation}
Both covariances are those of a linear Gaussian state-space model and follow from covariance recursions (Appendix~\ref{app:information}).
The one-step-ahead form of the mutual information is the directed information from the force to the data given the residuals \citep{massey1990causality},
\begin{equation}\label{eq:directed-information}
  \begin{aligned}
    D_\xi&=\sum_{k=1}^K I_\xi\big(U;y_k\mid y_{1:k-1},z_{1:n(k)}=0\big)\\
    &=\frac12\sum_{k=1}^K\log\frac{|\Sigma_k|}{|\tilde \Sigma_k|},
  \end{aligned}
\end{equation}
with $\Sigma_k$ the innovation covariance of $y_k$ in the filter that computes $\lobs$ (Appendix~\ref{app:update-equations}) and $\tilde \Sigma_k$ the same with the force known.

\paragraph{Diffusion sweeps.}
Figure~\ref{fig:force-information-duffing} and the sweeps of Appendix~\ref{app:state-diffusion} isolate the role of the diffusion: the force hyperparameters are held at the truth and $\sigma_x^2$ is varied over a log-spaced grid, separately on each grid density $\kappa$.
At every $\sigma_x^2$ the posterior is computed with the iterated extended Kalman smoother (IEKS) \citep{bell1994iterated}, and the force error and the criteria of Table~\ref{tab:criteria} are evaluated at the converged linearization (Appendix~\ref{app:sweeps}).
Each curve shows which diffusion a criterion would select if the force prior were known.
Calibrating the diffusion on the residual evidence $\lode$ treats it only as a numerical uncertainty and selects hard enforcement.
$I_\xi$ decreases towards both failure directions, and its maximizer lies on the plateau of low force error on every grid.

\begin{figure}[h!]
  \centering
  \includegraphics[width=\columnwidth]{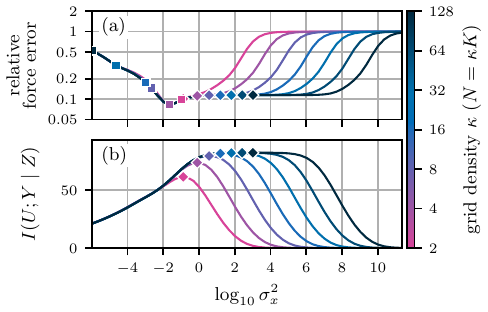}
  \caption{\textbf{The force error is smallest where $I(U;Y\mid Z)$ is largest.} (a) Relative force error and (b) mutual information $I(U;Y\mid Z)$ against the state diffusion $\sigma_x^2$ on Duffing (Appendix~\ref{app:state-diffusion}), with the IEKS and $\varphi_u$ at the truth. One line per grid density $\kappa$ (color bar). Per grid, \pickdiamond\ marks the maximizer of $I(U;Y\mid Z)$ and \picksquare\ that of $\lode$.}
  \label{fig:force-information-duffing}
\end{figure}

\subsection{Stable and Calibrated Latent Force ODE Filters}\label{subsec:calibration-guide}

Algorithm~\ref{alg:calibration-guide} implements the latent force ODE filter with calibration of the state diffusion and the force hyperparameters.
The state prior is chosen as smooth as the force allows, $q_x=m+q_u-1$, where $m$ is the order of the ODE.
A filter pass on a coarse initial grid can diverge, which the local error estimate of ODE filters detects \citep{bosch2021calibrated}, and the algorithm then moves to a finer grid (Appendix~\ref{app:stability}).
The posterior is computed with an IEKS, re-linearizing at the smoothed trajectory until convergence \citep{tronarp2021bayesian}.
Within each iteration the diffusion $\sigma_x^2$ is selected by a criterion $\mathcal J$ of Table~\ref{tab:criteria}, and the force parameters $\varphi_u$ are fitted on the observation evidence $\lobs$, both at the fixed trajectory $\xi$ and with L-BFGS-B \citep{byrd1995limited}.
Since $\varphi_u$ is fitted on $\lobs$ for every criterion, the variants differ only in how the diffusion is selected.
In the algorithm, $\mu^u_\kappa$ is the posterior mean of the force at the $K$ observation times on grid $\kappa$, and $s^u_\kappa$ its posterior standard deviation at the interval midpoints.
A grid is accepted once the force estimate has moved by less than $\tau_u$ posterior standard deviations from the previous grid and reproduces the data through the ODE, $r_\kappa\le\tau_r$.
Here $r_\kappa$ is the root-mean-square whitened residual of the observations against the ODE solved from the smoothed state with the force mean $\mu^u_\kappa$ as a known input (Appendix~\ref{app:acceptance}).

\begin{algorithm}[ht]
  \caption{Latent force ODE filter with the diffusion selected by a criterion $\mathcal J$ (Table~\ref{tab:criteria}).}
  \label{alg:calibration-guide}
  \begin{algorithmic}[1]
    \Require ($Y$, $m$, $q_u$, $\mathcal J$, $\varphi^{(0)}$, $\tau_\ell$, $\tau_u$, $\tau_r$, $n_{\max}$, $\kappa_{\max}$)
    \Ensure $p(X,U\mid Y,Z)$, $\varphi=(\sigma_x^2,\varphi_u)$
    \State $q_x\gets m+q_u-1$ \label{alg:qx}
    \State $\kappa_0\gets$ coarsest grid with stable EK1 at $\varphi^{(0)}$ \label{alg:kappa0}
    \State $\varphi\gets\varphi^{(0)}$,\; $\xi\gets$ predicted means of this EK1 pass \label{alg:init}
    \For{$\kappa=\kappa_0,2\kappa_0,\dots,\kappa_{\max}$} \label{alg:grid}
      \For{$i=1,\dots,n_{\max}$} \Comment{IEKS} \label{alg:inner-start}
        \State $\sigma_x^2\gets\argmax_{\sigma_x^2}\mathcal J^{\xi}$ \Comment{L-BFGS-B} \label{alg:fit}
        \State $\varphi_u\gets\argmax_{\varphi_u}\lobs^{\xi}$ \Comment{L-BFGS-B} \label{alg:fit-u}
        \State $\N(\mu,P)\gets p_\xi(X,U\mid Y,Z;\varphi)$ \Comment{smoother} \label{alg:smooth}
        \State $\xi\gets\mu$ \label{alg:relin}
        \State \textbf{if} $\lobs^{\xi}$ changed by $<\tau_\ell$ \textbf{then break} \label{alg:inner-end}
      \EndFor
      \State \textbf{if} $\|\mu^u_\kappa-\mu^u_{\kappa/2}\|<\tau_u\|s^u_\kappa\|$ \textbf{and} $r_\kappa\le\tau_r$ \textbf{then break} \label{alg:stop}
    \EndFor
  \end{algorithmic}
\end{algorithm}

\section{Experiments}\label{sec:experiments}

We run Algorithm~\ref{alg:calibration-guide} with each criterion $\mathcal J$ of Table~\ref{tab:criteria} and compare the criteria with each other and with the classical latent force model.
The main experiment (Sec.~\ref{subsec:setup}) uses three nonlinear systems driven by a force drawn from the model's prior, so that the only unknowns are the force and the hyperparameters that the algorithm calibrates.
Selecting the diffusion by $I(U;Y\mid Z)$ recovers the force on every problem (Sec.~\ref{subsec:d123}).
Section~\ref{subsec:off-model} repeats the comparison off-model, with forces outside the prior, details in Appendix~\ref{app:robustness}.

\subsection{Setup}\label{subsec:setup}

\paragraph{Data.}\label{subsec:systems}
Each of the three systems below is driven by one scalar force drawn from a Matérn-$5/2$ prior ($q_u=2$) and observed at $K$ equidistant times on $[0,T]$ under Gaussian noise of known variance.
Noise levels are reported as a signal-to-noise ratio (SNR) in amplitude, the standard deviation of the noise-free observed signal divided by the noise standard deviation.
For each system we draw force and noise $20$ times, which gives $60$ problems.

\paragraph{FitzHugh--Nagumo.}
\begin{equation}\label{eq:fhn}
  \begin{aligned}
    \dot v&=v-\tfrac13v^3-w+\bar I+u(t),\\
    \dot w&=\varepsilon\,(v+a-bw),
  \end{aligned}
\end{equation}
\citep{fitzhugh1961impulses,nagumo1962active} with $(a,b,\varepsilon)=(0.7,0.8,0.08)$, $\bar I=0.5$ and $(v,w)(0)=(-1,-0.5)$.
As in \citet{arnold2018approach}, only the membrane potential $v$ is observed, at $K=300$ times on $[0,100]$ with an SNR of $5$.
The force has $\sigma_u=0.35$ and $\ell_u=5$.

\paragraph{Duffing.}
\begin{equation}\label{eq:duffing}
  \ddot x=-x-2\zeta\dot x-\varepsilon x^3+u(t),
\end{equation}
\citep{kovacic2011duffing} with $\zeta=0.1$, $\varepsilon=0.5$ and $(x,\dot x)(0)=(0,0)$.
As in input-state estimation from accelerometers \citep{rogers2020bayesian}, the acceleration $\ddot x$ is observed at $K=128$ times on $[0,62.5]$ with an SNR of $100$.
The force has $\sigma_u=0.32$ and $\ell_u=3.2$.

\paragraph{Lorenz-96.}
\begin{equation}\label{eq:l96}
  \dot x_i=(x_{i+1}-x_{i-2})\,x_{i-1}-x_i+F+u(t)\,\delta_{i0},
\end{equation}
\citep{lorenz1996predictability} with cyclic indices, $i=1, \dots, D=8$ sites, $F=8$ and the initial state on the attractor.
The force acts on site $0$, and the sites $x_0,x_2,x_4,x_6$ are observed at $K=100$ times on $[0,10]$ with an SNR between $2.8$ and $4.5$ across the sites.
The force has $\sigma_u=2.5$ and $\ell_u=1$.

\paragraph{Model and baseline.}
Every criterion runs Algorithm~\ref{alg:calibration-guide} from the same initialization (Appendix~\ref{sec:initialization}).
The reported fit is the one on the grid that the algorithm accepts, or on the last grid if none is accepted (Appendix~\ref{app:acceptance}).
The classical latent force model runs through the same algorithm with the ODE filter replaced by a Runge--Kutta discretization treated as exact, so it has no diffusion to select, except on Lorenz-96, where an isotropic process noise on the state is fitted with $\varphi_u$ (Appendix~\ref{app:baselines}).

\paragraph{Metrics.}
We score the force $u$ and one state component (Duffing: $x$, FitzHugh--Nagumo: $v$, Lorenz-96: $x_0$) at the $n$ midpoints $t_i$ between observations.
For a scored quantity $g$ with posterior mean $\mu_g$ and variance $\sigma_g^2$, the errors are $e_i=g^\star(t_i)-\mu_g(t_i)$, the \textbf{relative error} is $\|e\|/\|g^\star\|$, and calibration is judged via the \textbf{Chi-square statistic} \citep{barshalom2001estimation}
\begin{equation*}
  \chi^2_g=\frac{1}{n}\sum_{i=1}^n \frac{e_i^2}{\sigma_g^2(t_i)}.
\end{equation*}
A calibrated posterior gives $\chi^2_g\approx1$. Lower and higher values imply under- and overconfidence, respectively.
The \textbf{error ratio} of a method on a problem is its relative force error divided by the smallest relative force error that any method reaches on that problem.
A method misses the force when its error ratio exceeds $1.5$.
We summarize the fitted force hyperparameters by the force roughness $\sigma_u^2/\ell_u^{2\nu}$ with $\nu=q_u+1/2$, the combination of the Matérn hyperparameters that data on a fixed time window can at best identify \citep{zhang2004inconsistent}.
The \textbf{roughness ratio} $(\sigma_u^2/\ell_u^{2\nu})/(\sigma_u^{\star 2}/\ell_u^{\star 2\nu})$ is the roughness of the fitted force prior divided by that of the true prior, below one for a force that is too smooth and above one for a force that is too rough.

\paragraph{Implementation.}
All methods are implemented in JAX \citep{jax2018github} in double precision and run single-threaded on the CPU of an Apple M4 Pro machine, for about $30$ CPU hours in total.
Appendix~\ref{app:implementation} gives the tolerances of Algorithm~\ref{alg:calibration-guide}, the initialization, the search boxes and the remaining implementation details.
All code is publicly available\footnote{\url{https://github.com/paufisch/lfcal}}.

\subsection{Results}\label{subsec:d123}

\begin{figure*}[t]
  \centering
  \includegraphics[width=\textwidth]{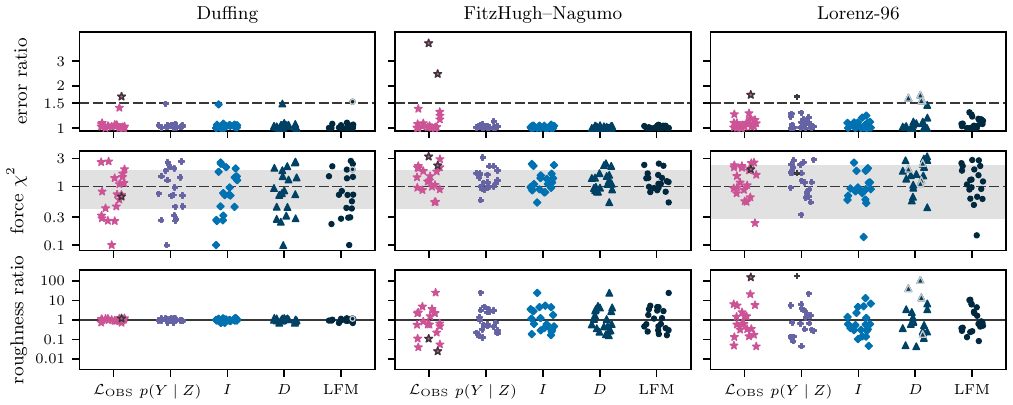}
  \caption{\textbf{Error ratio, force $\chi^2$ and roughness ratio of the criteria and the classical LFM on the $60$ problems.} Columns: Duffing, FitzHugh--Nagumo and Lorenz-96, one dot per problem and method. Top: error ratio, the relative force error over the smallest relative force error of any method on the same problem, with the miss threshold $1.5$ dashed. Fits above it are outlined, also in the rows below. Middle: force $\chi^2$, with the 95\,\% interval of a calibrated posterior whose errors are correlated as the true force prior in gray (Appendix~\ref{app:chi2}). Bottom: roughness ratio $(\sigma_u^2/\ell_u^{5})/(\sigma_u^{\star 2}/\ell_u^{\star 5})$ of the fitted force prior to the true one, below one for a force that is too smooth and above one for a force that is too rough.}
  \label{fig:comparison}
\end{figure*}

Figure~\ref{fig:comparison} compares the criteria on the $60$ problems by error ratio, force $\chi^2$ and roughness ratio, and Appendix~\ref{app:experiments} shows every fit that misses the force.
Every method recovers the force on at least $56$ of the $60$ problems, so the criteria differ in the few fits that fail.
Selecting the diffusion by $I(U;Y\mid Z)$ recovers the force on every problem and gives the best calibrated force bands.
$\lobs$ misses the force on $4$ problems (Fig.~\ref{fig:failures-lobs}).
On two of them, it selects a weakly enforced diffusion, at which the force is too smooth.
With $\varphi_u$ held at the truth, $\lobs$ is maximized at the weak-enforcement edge on these problems (Fig.~\ref{fig:force-information}), so they are failures of the criterion, not of its optimization.
$\log p(Y\mid Z)$ misses the force on $1$ problem, a Lorenz-96 problem that $\lobs$ also misses, with a force that is too rough (Fig.~\ref{fig:failures-logp-lfm}, left).
$D$ misses the force on $4$ problems of Lorenz-96, $3$ of them with a force that is too rough, and is otherwise as accurate as $I$ (Fig.~\ref{fig:failures-di}).
The classical model misses the force on a single problem, just above the threshold (Fig.~\ref{fig:failures-logp-lfm}, right).

A fit takes about a minute on a single CPU thread.
The median over the three systems ranges from $50$ to $72$\,s with $I(U;Y\mid Z)$, despite its four filter passes per evaluation (Table~\ref{tab:criteria}), from $42$ to $86$\,s with the other criteria, and from $37$ to $56$\,s with the classical model, whose grid ladder always runs to the end (Appendix~\ref{app:baselines}).

\subsection{Off-model data}\label{subsec:off-model}

We repeat the comparison with truth forces that are not draws of the model's prior, a rougher Matérn-$3/2$ draw, a sinusoid and a piecewise-constant force ($10$ problems per system and force), and on an SIR model in which the force sets the transmission rate and enters through the state ($20$ problems).
Appendix~\ref{app:robustness} gives the forces, the SIR model, the full results (Fig.~\ref{fig:robustness}) and example posteriors (Fig.~\ref{fig:robustness-examples}).
$I(U;Y\mid Z)$ misses the force on $1$ of these $110$ problems, against $12$ for $\lobs$, $6$ for $\log p(Y\mid Z)$, $4$ for $D$ and none for the classical model.
Most failures of the evidence criteria are on the rough forces, with overconfident bands, and those of $D$ on the sinusoid, with wide ones.
On the SIR model, where the force enters through the state, $I(U;Y\mid Z)$, $D$ and the classical model recover the force on every problem, the evidence criteria miss one each, and the force $\chi^2$ lies in the calibrated interval on nearly every problem.
For the piecewise-constant force the bands of every method are mostly overconfident, since a Matérn-$5/2$ prior is unsuitable to model jumps.

\section{Related Work}\label{sec:related-work}

\paragraph{Latent force models.}
LFMs were introduced for linear ODEs, where a GP prior on the force induces a GP prior on the state through the Green's function \citep{alvarez2009latent}, and nonlinearity remains their main limitation.
In state-space form \citep{hartikainen2010kalman}, inference is linear in time \citep{hartikainen2011sequential}, approximate Gaussian filtering reaches nonlinear dynamics \citep{hartikainen2012state,sarkka2023bayesian}, and LFMs enter stochastic and model predictive control \citep{sarkka2018gaussian,landgraf2022nonlinear}.
Beyond filtering, nonlinear LFMs use variational inference with normalizing flows \citep{ward2020black}, adjoints through a numerical solver \citep{moss2021approximate} and deep GPs of ODE-based process convolutions \citep{baldwin2025deep}.

\paragraph{Input estimation in structural dynamics.}
Structural dynamics treats the same problem as input-state or input-state-parameter estimation, mostly by appending the force to the state of a Kalman filter, as in the augmented (AKF) \citep{lourens2012augmented,naets2015online} and dual Kalman filter \citep{azam2015dual}.
\citet{nayek2019gaussian} bring GP latent force models (GPLFMs) to the field and show that AKF variants are GPLFMs with random-walk priors, and \citet{rogers2020application} add parameter estimation.
The unbiased minimum-variance filter \citep{gillijns2007unbiased} is the infinite-variance limit of the AKF \citep{ding2023infinity}, with a bias-variance tradeoff in between \citep{li2026bias}.

\paragraph{Calibrating ODE filters on data.}
ODE filters conditioned on measurements have mostly served parameter inference of the vector field.
\citet{kersting2020differentiable} include the data in the filter pass and fit the parameters to the resulting likelihood by gradient descent.
\citet{tronarp2022fenrir} condition on the ODE first and score the data under the resulting physics-enhanced prior, optimizing the diffusion jointly with the parameters.
\citet{beck2024diffusion} read this diffusion as a weight of the data against the ODE solution and lower it along a fixed schedule.
\citet{wu2024data} compute $p(Y\mid Z)=p(Y,Z)/p(Z)$ from two filter passes, again fitting the prior scales jointly with the parameters.
In latent force inference, by contrast, \citet{schmidt2021probabilistic} fix the diffusion per experiment, and its role has not been analyzed.
Sec.~\ref{subsec:posterior-factorization} does so and explains why the residual evidence selects the wrong direction.
The conditional evidence $\log p(Y\mid Z)$ of Table~\ref{tab:criteria} is the likelihood of \citet{tronarp2022fenrir} and \citet{wu2024data}, and $\lobs$ its one-step-ahead form (Sec.~\ref{subsec:info-criterion}).

\section{Limitations}
The experiments use equidistant measurements and fixed grids without adaptive steps, so stiff dynamics may need finer grids than Algorithm~\ref{alg:calibration-guide} reaches.
The filter scales cubically in the dimension of the extended state, which limits the method to low-dimensional systems.
Nonlinear dynamics are handled by the Gaussian approximation of the extended Kalman filter, and beyond the single update of Sec.~\ref{subsec:state-diffusion} the results of this paper are empirical findings without guarantees.
The classical latent force model is competitive on every system here, and a calibrated latent force ODE filter matches rather than surpasses it.

\section{Conclusion}
This work augments the latent force ODE filter framework, which was presented by \citet{schmidt2021probabilistic} mainly in the context of experimental evaluation, with a theoretical analysis of the resulting joint posterior over state and latent force.
Besides a better understanding of the relationship of the method to classic latent force models, the findings entail best practices for model selection by surfacing a crucial dual role of the prior state diffusion parameter.
In doing so, this paper advances the framework towards broad applicability for a variety of real-world use cases.

\subsubsection*{Acknowledgments}
The authors gratefully acknowledge financial support by the European Research Council through ERC CoG Action 101123955 ANUBIS; the DFG Cluster of Excellence ``Machine Learning -- New Perspectives for Science'', EXC 2064/2; the DFG SPP 2298 (Project HE 7114/6-1), and the Carl Zeiss Foundation, (project ``Certification and Foundations of Safe Machine Learning Systems in Healthcare''), as well as funds from the Ministry of Science, Research and Arts of the State of Baden-W\"urttemberg.
The authors thank the International Max Planck Research School for Intelligent Systems (IMPRS-IS) for supporting Paul Fischer.

{
\small
\bibliographystyle{apalike}
\bibliography{bibliography}

}

\clearpage
\onecolumn
\aistatstitle{Appendix}
\appendix
\counterwithin{figure}{section}
\counterwithin{table}{section}

\section{Additional experimental results}\label{app:experiments}

\subsection{Failures}

\begin{figure}[ht]
  \centering
  \includegraphics[width=\textwidth]{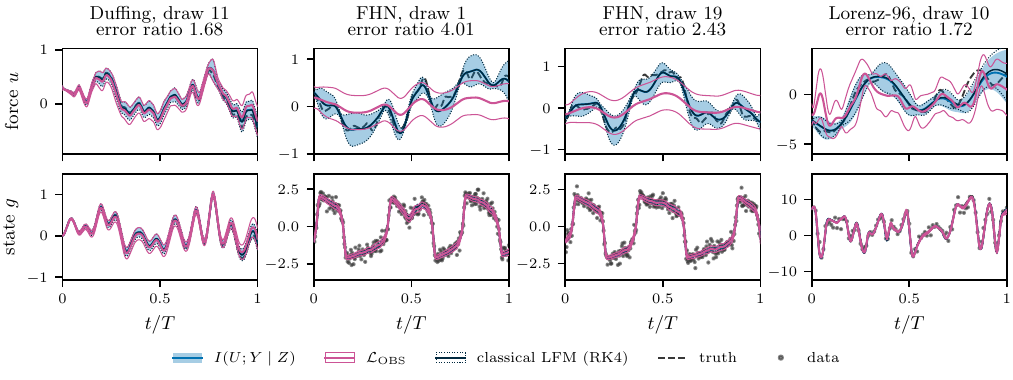}
  \caption{\textbf{The $4$ misses of $\lobs$.} Force (top) and the scored state with its data (bottom) against $t/T$. On Duffing the acceleration is observed, so no data are shown. Bands: mean $\pm 2$ standard deviations. Purple (outline): $\sigma_x^2$ by $\lobs$. Blue: $\sigma_x^2$ by $I(U;Y\mid Z)$. Navy (dotted): classical LFM. Dashed: truth.}
  \label{fig:failures-lobs}
\end{figure}

\begin{figure}[ht]
  \centering
  \includegraphics[width=\textwidth]{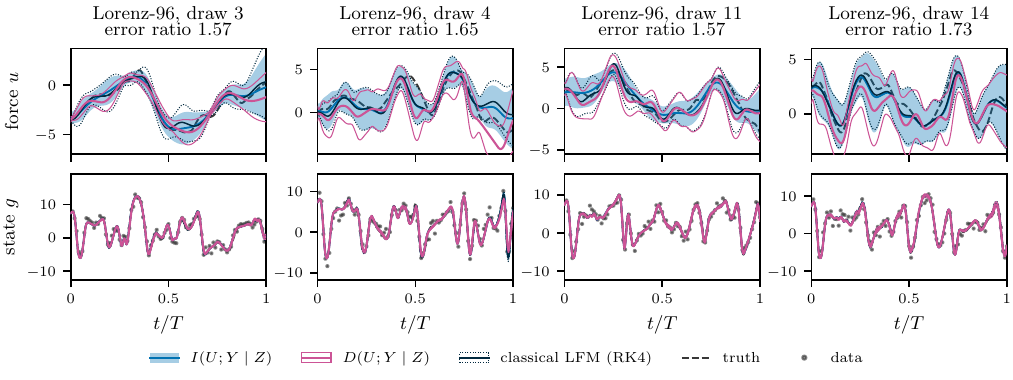}
  \caption{\textbf{The $4$ misses of $D$, all on Lorenz-96.} Force (top) and the observed site $x_0$ with its data (bottom) against $t/T$. Bands: mean $\pm 2$ standard deviations. Purple (outline): $\sigma_x^2$ by $D(U;Y\mid Z)$. Blue: $\sigma_x^2$ by $I(U;Y\mid Z)$. Navy (dotted): classical LFM. Dashed: truth.}
  \label{fig:failures-di}
\end{figure}

\begin{figure}[ht]
  \centering
  \includegraphics[width=\textwidth]{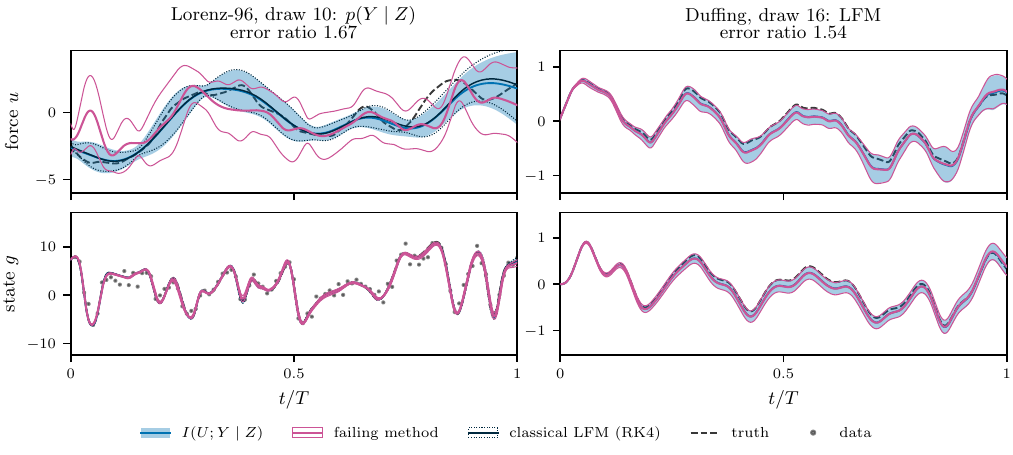}
  \caption{\textbf{The miss of $\log p(Y\mid Z)$ (left) and of the classical model (right).} Force (top) and the scored state (bottom) against $t/T$. Bands: mean $\pm 2$ standard deviations. Purple (outline): the failing method. Blue: $\sigma_x^2$ by $I(U;Y\mid Z)$. Navy (dotted): classical LFM. Dashed: truth.}
  \label{fig:failures-logp-lfm}
\end{figure}

Figures~\ref{fig:failures-lobs}, \ref{fig:failures-di} and~\ref{fig:failures-logp-lfm} show every fit that misses the force, with an error ratio above $1.5$, on the $60$ problems of Sec.~\ref{subsec:setup}, one column per problem, with the fit of $I(U;Y\mid Z)$ and of the classical model on the same problem.

\subsection{Forces outside the prior and the SIR model}\label{app:robustness}

The misspecified forces keep the model of Sec.~\ref{subsec:systems}, a Matérn-$5/2$ prior with unknown $\varphi_u$, but draw the truth differently, each with the marginal standard deviation $\sigma_u$ of the prior.
The Matérn-$3/2$ force is a draw of the rougher prior with the same length scale.
The sinusoid has the angular frequency $\omega\,\ell_u^{-1}$ with $\omega$ uniform on $[0.7,1.4]$ and a uniform phase.
The piecewise-constant force switches at the times of a Poisson process with mean spacing $2\ell_u$, to levels drawn from $\N(0,\sigma_u^2)$.
The SIR model \citep{kermack1927contribution} describes the susceptible and infected fractions of a population under a time-varying transmission rate,
\begin{equation*}
  \dot S=-\beta_0 e^{u(t)} S I,\qquad \dot I=\beta_0 e^{u(t)} S I-\gamma I,
\end{equation*}
with $\beta_0=0.15$, $\gamma=0.1$ per day and $(S,I)(0)=(0.999,0.001)$, and the recovered fraction $R=1-S-I$ left implicit.
The force enters multiplicatively through the state, so the ODE is not affine in the force.
The infected fraction is observed daily on $[0,150]$, $K=150$, with an SNR of $10$.
The force has $\sigma_u=0.25$ and $\ell_u=10$ days.

\begin{figure}[ht]
  \centering
  \includegraphics[width=\textwidth]{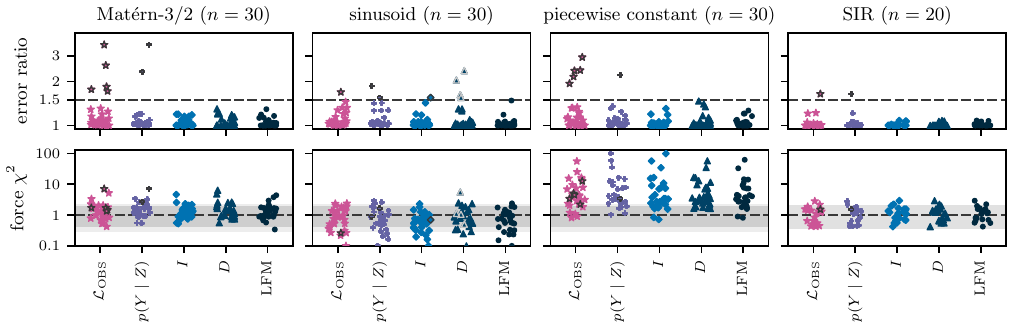}
  \caption{\textbf{Error ratio and force $\chi^2$ off the model.} One dot per problem and method, for truth forces outside the model's prior (pooled over Duffing, FitzHugh--Nagumo and Lorenz-96, $10$ problems each) and for the SIR model. Top: error ratio, with the miss threshold $1.5$ dashed. Fits above it are outlined, also in the row below. Bottom: force $\chi^2$, with the 95\,\% interval of a calibrated posterior whose errors are correlated as the model's force prior in gray (Appendix~\ref{app:chi2}), one interval per pooled system, drawn on top of each other.}
  \label{fig:robustness}
\end{figure}

\begin{figure}[ht]
  \centering
  \includegraphics[width=\textwidth]{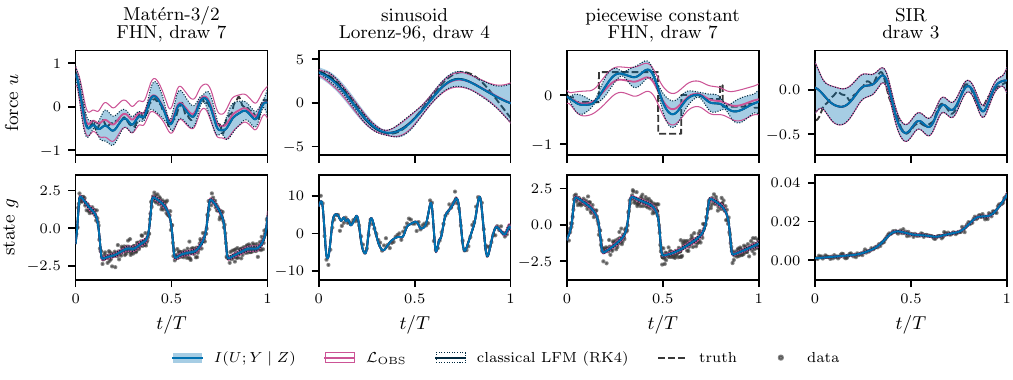}
  \caption{\textbf{Example posteriors off the model.} One problem per kind of truth force (columns), chosen as in the text. Force (top) and the scored state with its data (bottom) against $t/T$. Bands: mean $\pm 2$ standard deviations. Blue: $\sigma_x^2$ by $I(U;Y\mid Z)$. Purple (outline): $\sigma_x^2$ by $\lobs$. Navy (dotted): classical LFM. Dashed: truth. The posteriors nearly coincide, and for the piecewise-constant force all of them round off the jumps, which a Matérn-$5/2$ prior cannot represent.}
  \label{fig:robustness-examples}
\end{figure}

Figure~\ref{fig:robustness} compares the criteria on all $110$ problems.
Figure~\ref{fig:robustness-examples} shows a typical problem of each kind on which every method has an error ratio below $1.5$, the one on which the force error of $I(U;Y\mid Z)$ is closest to its median over such problems.

\section{Matérn priors in state-space form}\label{bsec:matern}

A stationary Gaussian process with Matérn-$\nu$ covariance $k(\tau) \propto (\lambda |\tau|)^\nu K_\nu(\lambda|\tau|)$, where $K_\nu$ is the modified Bessel function of the second kind and $\lambda > 0$ is an inverse length scale parameter, admits a state-space representation whenever $\nu = q + 1/2$ for $q \in \{0, 1, 2, \dots\}$ \citep{hartikainen2010kalman, sarkka2019applied}.
The corresponding SDE is a stable linear system of order $q+1$ with transfer function from the driving noise to the process value
\begin{equation*}
  H_{\text{Mat\'ern}}(s)
  = \frac{\sigma_u \sqrt{2\nu/\ell_u^2}^{\,\nu}}{(s + \lambda)^{\nu + 1/2}},
  \qquad
  \lambda = \sqrt{\frac{2\nu}{\ell_u^2}},
\end{equation*}
with length scale $\ell_u > 0$ and marginal variance $\sigma_u^2$.
Its power spectral density $|H(\mathrm{i}\omega)|^2 \propto (\lambda^2 + \omega^2)^{-(\nu + 1/2)}$ is flat for $|\omega| \ll \lambda$ and rolls off as $\omega^{-(2\nu+1)}$ above the cutoff $\lambda$, which is set by the length scale.
The smoothness $\nu$ sets the sample path regularity: Matérn-$\nu$ sample paths are $\lceil\nu\rceil - 1$ times differentiable, and the state vector carries the process and its first $q$ derivatives \citep{hartikainen2010kalman}.

\section{Update equations}\label{app:update-equations}

The prediction and update equations of the latent force ODE filter are presented here in more detail and in block form.
Both ODE and observation updates are performed using the extended Kalman filter update \citep{sarkka2023bayesian}.

\paragraph{State} At the previous step, $p\big(\tilde{X}\big) = \N (\mu, P)$:
\begin{equation*}
  \mu = \begin{bmatrix} \mu_x \\ \mu_u \end{bmatrix},
  \quad
  P = \begin{bmatrix} P_{xx} & P_{xu} \\ P_{ux} & P_{uu} \end{bmatrix},
  \quad P_{ux} = P_{xu}^{\top}.
\end{equation*}

\paragraph{Predict} One step ahead,
$p(\tilde{X}) = \N(\pre{\mu}, \pre{P}) = \N(\tilde{A}_h \mu, \tilde{A}_h P \tilde{A}_h^{\top} + \tilde{Q}_h)$:
\begin{equation*}
  \begin{aligned}
  \pre{P_{xx}} &= A_h P_{xx}A_h^{\top} + Q_h,\\
  \pre{P_{xu}} &= A_h P_{xu}\bar{A}_h^{\top},\\
  \pre{P_{uu}} &= \bar{A}_h P_{uu}\bar{A}_h^{\top} + \bar{Q}_h.
  \end{aligned}
\end{equation*}

\paragraph{Update (ODE)} $p(\tilde{X} \mid z) = \N(\ode{\mu}, \ode{P}) = \N(\pre{\mu} - L\hat{z}, \pre{P} - LSL^{\top})$:

The residual $\mathcal{Z}[\tilde{X}]$ is replaced by its first-order Taylor expansion at the predicted mean (extended Kalman filter),
\begin{equation*}
  \mathcal{Z}[\tilde{X}] \approx C\,(X - \pre{\mu_x}) + B\,(U - \pre{\mu_u}) + \hat{z},
\end{equation*}
with predicted residual $\hat{z} = \mathcal{Z}[\pre{\mu}]$ and Jacobians
\begin{equation*}
  C = \frac{\partial \mathcal{Z}}{\partial X}\bigg\rvert_{\pre{\mu}},
  \qquad
  B = \frac{\partial \mathcal{Z}}{\partial U}\bigg\rvert_{\pre{\mu}} = -\frac{\partial f}{\partial U}\bigg\rvert_{\pre{\mu}}.
\end{equation*}
Innovation covariance, with $B$ and $C$ the blocks at the current node of the stacked Jacobians of Sec.~\ref{subsec:posterior-factorization}:
\begin{equation*}
  S = \underbrace{C\pre{P_{xx}}C^{\top}}_{S_{x}}
    + \underbrace{B\pre{P_{uu}}B^{\top}}_{S_{u}}
    + \underbrace{\big(C\pre{P_{xu}}B^{\top} + B\pre{P_{ux}}C^{\top}\big)}_{2S_{xu}}.
\end{equation*}
Kalman gain:
\begin{equation*}
  \begin{aligned}
    \begin{bmatrix} L_x \\ L_u \end{bmatrix}
     &= \begin{bmatrix}\big(\pre{P_{xx}}C^{\top} + \pre{P_{xu}}B^{\top}\big)S^{-1}\\
        \big(\pre{P_{ux}}C^{\top} + \pre{P_{uu}}B^{\top}\big)S^{-1}\end{bmatrix}.
  \end{aligned}
\end{equation*}
Posterior covariance:
\begin{equation*}
  \begin{aligned}
    \ode{P_{xx}} &= \pre{P_{xx}} - L_xSL_x^{\top},\\
    \ode{P_{xu}} &= \pre{P_{xu}} - L_xSL_u^{\top}, \\
    \ode{P_{uu}} &= \pre{P_{uu}} - L_uSL_u^{\top}.
  \end{aligned}
\end{equation*}

\paragraph{Update (OBS)} $p(\tilde{X} \mid z, y) = \N(\mu, P) = \N(\ode{\mu} + G\,(y - \hat y), \ode{P} - G\Sigma G^{\top})$:

The observation $h[X]$ is expanded to first order at the ODE-updated mean, $h[X] \approx h[\ode{\mu_x}] + H(X - \ode{\mu_x})$ with $H = \partial h/\partial X\rvert_{\ode{\mu}}$ (exact for linear $h$), giving the joint observation operator $\tilde H=[H\;\;0]$ and predicted observation $\hat y = h[\ode{\mu_x}]$.
Innovation covariance:
\begin{equation*}
  \Sigma = H\ode{P_{xx}} H^{\top} + \Robs.
\end{equation*}
Kalman gain:
\begin{equation*}
    G = \begin{bmatrix}
      G_x\\G_u
    \end{bmatrix} =
  \begin{bmatrix} \ode{P_{xx}}H^{\top}\Sigma^{-1} \\ \ode{P_{ux}}H^{\top}\Sigma^{-1} \end{bmatrix}.
\end{equation*}
Posterior covariance:
\begin{equation*}
  \begin{aligned}
    P_{xx} &= \ode{P_{xx}} - G_x\Sigma G_x^{\top},\\
    P_{xu} &= \ode{P_{xu}} - G_x\Sigma G_u^{\top},\\
    P_{uu} &= \ode{P_{uu}} - G_u\Sigma G_u^{\top}.
  \end{aligned}
\end{equation*}

\subsection{Single update from an independent prior}\label{app:single-update}

This subsection writes out one residual update and one observation update for a prediction in which state and force are independent, and makes every dependence on the state diffusion $\sigma_x^2$ explicit.
It is the example of Secs.~\ref{subsec:state-diffusion} and~\ref{subsec:failure-directions}.

\paragraph{Prior} The prediction has zero mean and no cross-covariance, and the diffusion is factored out of the state covariance, $\pre{P_{xx}}=\sigma_x^2\pre{\check P_{xx}}$, as in $Q_h=\sigma_x^2\check Q_h$:
\begin{equation*}
  p(X,U) = \N\!\left(\begin{bmatrix} X \\ U \end{bmatrix} \,\middle|\, \pre{\mu}, \pre{P}\right),
  \qquad
  \pre{\mu} = \begin{bmatrix} 0 \\ 0 \end{bmatrix},
  \qquad
  \pre{P} = \begin{bmatrix} \sigma_x^2\pre{\check P_{xx}} & 0 \\ 0 & \pre{P_{uu}} \end{bmatrix}.
\end{equation*}
The linearized residual is then $\mathcal{Z}[X,U] = CX + BU + \hat z$.

\paragraph{Update (ODE)} $p(X,U\mid z=0) = \N(\ode{\mu},\ode{P})$:

\begin{equation*}
  S = S_x + S_u,
  \qquad S_x = C\pre{P_{xx}}C^\top = \sigma_x^2\check S_x,
  \qquad \check S_x = C\pre{\check P_{xx}}C^\top,
  \qquad S_u = B\pre{P_{uu}}B^\top.
\end{equation*}
Kalman gain:
\begin{equation*}
  L_x = \pre{P_{xx}}C^\top S^{-1},
  \qquad
  L_u = \pre{P_{uu}}B^\top S^{-1}.
\end{equation*}
Posterior mean and covariance, with $SL_x^\top = C\pre{P_{xx}}$ and $SL_u^\top = B\pre{P_{uu}}$ inserted into $\pre{P}-LSL^\top$:
\begin{equation*}
  \ode{\mu} = -\begin{bmatrix} L_x \\ L_u \end{bmatrix}\hat z,
  \qquad
  \ode{P} = \begin{bmatrix} \ode{P_{xx}} & \ode{P_{xu}} \\ \ode{P_{ux}} & \ode{P_{uu}} \end{bmatrix}
  = \begin{bmatrix}
      \pre{P_{xx}} - L_x\,C\pre{P_{xx}} & -L_x\,B\pre{P_{uu}} \\
      -\pre{P_{uu}}B^\top L_x^\top & \pre{P_{uu}} - L_u\,B\pre{P_{uu}}
    \end{bmatrix}.
\end{equation*}

\paragraph{Force marginal and state conditional} The posterior factorizes as $p(X,U\mid z=0) = p(X\mid U,z=0)\,p(U\mid z=0)$.
Marginalizing the state out of the residual model gives the likelihood of the force,
\begin{equation*}
  p(z\mid U) = \N\big(z \mid BU+\hat z,\ S_x\big),
\end{equation*}
so that the force marginal is the force prior times a Gibbs factor with temperature $\sigma_x^2$:
\begin{equation*}
  p(U\mid z=0) = \N\big(U \mid \ode{\mu_u}, \ode{P_{uu}}\big) \propto p(U)\,\exp\!\left(-\frac{\mathcal E(U)}{2\sigma_x^2}\right),
  \qquad
  \mathcal E(U) = (BU+\hat z)^\top\check S_x^{-1}(BU+\hat z).
\end{equation*}
In precision form, $(\ode{P_{uu}})^{-1} = (\pre{P_{uu}})^{-1} + B^\top S_x^{-1}B$.
The ratio $\mathcal V(U;z) = p(U\mid z=0)/p(U)$ of Sec.~\ref{subsec:posterior-factorization} is this Gibbs factor up to a constant, and so is the weight $\mathcal W(U)$.
Given the force, the state is distributed as
\begin{equation*}
  p(X\mid U,z=0) = \N\big(X \mid -L_{x\mid u}(BU+\hat z),\ \ode{P_{x\mid u}}\big),
  \qquad
  L_{x\mid u} = \pre{P_{xx}}C^\top S_x^{-1},
  \qquad
  \ode{P_{x\mid u}} = \pre{P_{xx}} - L_{x\mid u}S_xL_{x\mid u}^\top.
\end{equation*}
The diffusion cancels in the gain, since $\pre{P_{xx}}$ and $S_x$ are both proportional to it: with the force given and $\Rode=0$, the state is the only source of uncertainty in the residual.
The mean of the state given the force therefore does not depend on the diffusion, and its covariance $\ode{P_{x\mid u}}$ is proportional to it.
The blocks of $\ode{P}$ decompose accordingly:
\begin{equation*}
  \ode{P_{xx}} = \ode{P_{x\mid u}} + L_{x\mid u}\,B\ode{P_{uu}}B^\top L_{x\mid u}^\top,
  \qquad
  \ode{P_{xu}} = -L_{x\mid u}\,B\ode{P_{uu}},
  \qquad
  B\ode{P_{uu}}B^\top = S_xS^{-1}S_u.
\end{equation*}

\paragraph{Update (OBS)} $p(X,U\mid z=0,y) = \N(\mu,P)$:

For a linear observation $y = HX+\epsilon$ with $\epsilon\sim\N(0,\Robs)$, the predicted observation is $\hat y = H\ode{\mu_x} = -HL_x\,\hat z$.
Innovation covariance and Kalman gain inherit the dependence on the diffusion from $\ode{P}$:
\begin{equation*}
  \Sigma = H\ode{P_{xx}}H^\top + \Robs,
  \qquad
  G_x = \ode{P_{xx}}H^\top\Sigma^{-1},
  \qquad
  G_u = \ode{P_{ux}}H^\top\Sigma^{-1}.
\end{equation*}
Posterior mean and covariance, with $\Sigma G_x^\top = H\ode{P_{xx}}$ and $\Sigma G_u^\top = H\ode{P_{xu}}$ inserted into $\ode{P}-G\Sigma G^\top$:
\begin{equation*}
  \mu = \ode{\mu} + \begin{bmatrix} G_x \\ G_u \end{bmatrix}(y-\hat y),
  \qquad
  P = \begin{bmatrix} P_{xx} & P_{xu} \\ P_{ux} & P_{uu} \end{bmatrix}
  = \begin{bmatrix}
      \big(I - G_xH\big)\ode{P_{xx}} & \big(I - G_xH\big)\ode{P_{xu}} \\
      \ode{P_{ux}}\big(I - G_xH\big)^\top & \ode{P_{uu}} - G_u\,H\ode{P_{xu}}
    \end{bmatrix}.
\end{equation*}

\paragraph{Limits of the diffusion}
Table~\ref{tab:diffusion-limits} lists the leading order of the quantities of both updates as $\sigma_x^2\to0$ and as $\sigma_x^2\to\infty$.
It assumes that $S_x$, $\Robs$ and $H\ode{P_{x\mid u}}H^\top$ are invertible.
$S_x$ is, since $\pre{P_{xx}}$ is positive definite after a prediction step and every row of $C$ contains the highest derivative $x^{(m)}$.
$S_u$ need not be: a scalar force that enters one of $d_x$ components of the ODE gives $\operatorname{rank}S_u=1$, as on FitzHugh--Nagumo, Lorenz-96 and the SIR model.
The limits for $\sigma_x^2\to0$ therefore use the pseudo-inverse $S_u^+$ and the projection onto the range of $B$ that is orthogonal in the metric $S_x^{-1}$,
\begin{equation*}
  \Pi = B\big(B^\top S_x^{-1}B\big)^+B^\top S_x^{-1},
\end{equation*}
which does not depend on the diffusion.
If the force reaches every component of the residual, as on Duffing, $\Pi=I$ and $S_u^+=S_u^{-1}$.

For $\sigma_x^2\to0$, the weight concentrates on the forces of least energy, $BU=-\Pi\hat z$.
The force takes up the part $\Pi\hat z$ of the residual that it can reach and is certain of it, $B\ode{P_{uu}}B^\top\to0$.
The state takes up the rest, $\ode{\mu_x}\to-L_{x\mid u}(I-\Pi)\hat z$, which vanishes if $\Pi=I$, and its covariance vanishes.
Both gains of the observation update vanish with $\sigma_x^2$, so the data change neither state nor force: to leading order, the moments after the observation update are those before it.

For $\sigma_x^2\to\infty$, the force keeps its prior, and the state covariance diverges.
The cross-covariance stays bounded: by the decomposition above, $\ode{P_{xu}} = -L_{x\mid u}B\,\ode{P_{uu}}$, where $L_{x\mid u}$ does not depend on the diffusion and $\ode{P_{uu}}$ is at most the prior variance of the force.
The state gain tends to $G_x^\infty = \ode{P_{x\mid u}}H^\top(H\ode{P_{x\mid u}}H^\top)^{-1}$ with $HG_x^\infty = I$, so the state interpolates the observation, $H\mu_x\to y$, and the remaining state covariance $(I - G_x^\infty H)\ode{P_{x\mid u}}$ is proportional to the diffusion.
It diverges in every component that neither the residual nor the observation fixes, such as the higher derivatives of an IWP($q_x$) state with $q_x\ge2$. In the multi-step filter these are the directions in which the state moves freely between the observations, where the bands of Fig.~\ref{fig:failure-directions} (right) widen.
The force gain vanishes as $\sigma_x^{-2}$, since $\ode{P_{ux}}$ stays bounded while $\Sigma$ diverges, so the force keeps its prior after the observation update as well.

\begin{table}[h]
  \centering
  \caption{\textbf{Limits of the single update as $\sigma_x^2\to0$ and $\sigma_x^2\to\infty$.} Leading order of the single residual update (top), of the innovation covariance and gains of the subsequent observation update (middle), and of the posterior after both updates (bottom), as $\sigma_x^2\to0$ and as $\sigma_x^2\to\infty$. \tozero{Blue}: the entry vanishes in the limit. \toinfty{Orange}: it diverges in proportion to $\sigma_x^2$. Black: it tends to a finite limit that does not depend on the diffusion. $\pre{P_{xx}}$, $S_x$ and $\ode{P_{x\mid u}}$ are proportional to $\sigma_x^2$. If the force reaches every component of the residual, $\Pi=I$, and $L_x$ and the state means vanish in proportion to $\sigma_x^2$ for $\sigma_x^2\to0$.}
  \label{tab:diffusion-limits}
  \renewcommand{\arraystretch}{1.25}
  \begin{tabular}{@{}l|ll@{}}
    \toprule
    & $\sigma_x^2\to0$ & $\sigma_x^2\to\infty$\\
    \midrule
    $L_x$ & $L_{x\mid u}(I-\Pi)$ & $L_{x\mid u}$\\
    $L_u$ & $\pre{P_{uu}}B^\top S_u^+\Pi$ & \tozero{$\pre{P_{uu}}B^\top S_x^{-1}\to0$}\\
    $\ode{\mu_x}$ & $-L_{x\mid u}(I-\Pi)\hat z$ & $-L_{x\mid u}\hat z$\\
    $\ode{\mu_u}$ & $-\pre{P_{uu}}B^\top S_u^+\Pi\hat z$ & \tozero{$0$}\\
    $\ode{P_{xx}}$ & \tozero{$\pre{P_{xx}}-L_{x\mid u}(I-\Pi)C\pre{P_{xx}}\to0$} & \toinfty{$\ode{P_{x\mid u}}\to\infty$}\\
    $\ode{P_{xu}}$ & \tozero{$-\pre{P_{xx}}C^\top\Pi^\top S_u^+B\pre{P_{uu}}\to0$} & $-L_{x\mid u}B\pre{P_{uu}}$\\
    $\ode{P_{uu}}$ & $\pre{P_{uu}}-\pre{P_{uu}}B^\top S_u^+B\pre{P_{uu}}$ & $\pre{P_{uu}}$\\
    \midrule
    $\Sigma$ & $\Robs$ & \toinfty{$H\ode{P_{x\mid u}}H^\top\to\infty$}\\
    $G_x$ & \tozero{$\ode{P_{xx}}H^\top(\Robs)^{-1}\to0$} & $G_x^\infty = \ode{P_{x\mid u}}H^\top(H\ode{P_{x\mid u}}H^\top)^{-1}$\\
    $G_u$ & \tozero{$-\pre{P_{uu}}B^\top S_u^+\Pi C\pre{P_{xx}}H^\top(\Robs)^{-1}\to0$} & \tozero{$-\pre{P_{uu}}B^\top L_{x\mid u}^\top H^\top(H\ode{P_{x\mid u}}H^\top)^{-1}\to0$}\\
    \midrule
    $\mu_x$ & $-L_{x\mid u}(I-\Pi)\hat z$ & $-L_{x\mid u}\hat z + G_x^\infty(y + HL_{x\mid u}\hat z)$\\
    $\mu_u$ & $-\pre{P_{uu}}B^\top S_u^+\Pi\hat z$ & \tozero{$0$}\\
    $P_{xx}$ & \tozero{$\pre{P_{xx}}-L_{x\mid u}(I-\Pi)C\pre{P_{xx}}\to0$} & \toinfty{$(I-G_x^\infty H)\ode{P_{x\mid u}}\to\infty$}\\
    $P_{xu}$ & \tozero{$-\pre{P_{xx}}C^\top\Pi^\top S_u^+B\pre{P_{uu}}\to0$} & $-(I-G_x^\infty H)L_{x\mid u}B\pre{P_{uu}}$\\
    $P_{uu}$ & $\pre{P_{uu}}-\pre{P_{uu}}B^\top S_u^+B\pre{P_{uu}}$ & $\pre{P_{uu}}$\\
    \bottomrule
  \end{tabular}
\end{table}

\paragraph{Mutual information}
For the single update, the two covariances of Eq.~\eqref{eq:mutual-information-gaussian} are the innovation covariance of the observation with the force unknown and with the force known,
\begin{equation*}
  \Sigma = H\ode{P_{xx}}H^\top + \Robs, \qquad \tilde\Sigma = H\ode{P_{x\mid u}}H^\top + \Robs.
\end{equation*}
By the decomposition of $\ode{P_{xx}}$ above, $\Sigma = \tilde\Sigma + HL_{x\mid u}\,B\ode{P_{uu}}B^\top L_{x\mid u}^\top H^\top$, so that
\begin{equation*}
  I(U;Y\mid Z) = \frac12\log\frac{|\Sigma|}{|\tilde\Sigma|}
  = \frac12\log\Big|I + \tilde\Sigma^{-1}HL_{x\mid u}\,B\ode{P_{uu}}B^\top L_{x\mid u}^\top H^\top\Big|.
\end{equation*}
The information is that of a signal in noise.
The signal is the variance the residual leaves in the force, $B\ode{P_{uu}}B^\top$, seen through the state: it vanishes as $S_x\Pi^\top=B(B^\top S_x^{-1}B)^+B^\top$ for $\sigma_x^2\to0$ and saturates at its prior value $S_u$ for $\sigma_x^2\to\infty$.
The noise $\tilde\Sigma$ is the observation noise plus the numerical uncertainty of the state given the force: it tends to $\Robs$ for $\sigma_x^2\to0$ and grows in proportion to $\sigma_x^2$ for $\sigma_x^2\to\infty$.
To leading order,
\begin{equation*}
  I(U;Y\mid Z) \approx
  \begin{cases}
    \tfrac12\operatorname{tr}\!\Big((\Robs)^{-1}HL_{x\mid u}S_x\Pi^\top L_{x\mid u}^\top H^\top\Big) \propto \sigma_x^{2}, & \sigma_x^2\to0,\\[1ex]
    \tfrac12\operatorname{tr}\!\Big((H\ode{P_{x\mid u}}H^\top)^{-1}HL_{x\mid u}S_uL_{x\mid u}^\top H^\top\Big) \propto \sigma_x^{-2}, & \sigma_x^2\to\infty.
  \end{cases}
\end{equation*}
The information thus vanishes in both failure directions and is largest in between, as in Fig.~\ref{fig:force-information-duffing}.

\subsection{Covariance recursions for the information criteria}\label{app:information}

At a frozen linearization the model is linear and Gaussian, and the covariances of Eqs.~\eqref{eq:mutual-information-gaussian} and~\eqref{eq:directed-information} do not depend on the values of the data.
Their log-determinants follow from the prediction error decomposition: in a Kalman filter that processes the residuals $z_n$ and the observations $y_k$ in time order, the observation after the residual at the same node, $\log|\operatorname{Cov}(Z,Y)|$ is the sum of the log-determinants of all innovation covariances.
Only the covariance part of the filter of Appendix~\ref{app:update-equations} is needed.
With the joint residual Jacobian $\tilde C_n=[C_n\;\;B_n]$ at node $n$,
\begin{equation*}
  \pre{P}_n=\tilde A_hP_{n-1}\tilde A_h^\top+\tilde Q_h,
  \qquad
  S_n=\tilde C_n\pre{P}_n\tilde C_n^\top,
  \qquad
  \ode{P}_n=\pre{P}_n-\pre{P}_n\tilde C_n^\top S_n^{-1}\tilde C_n\pre{P}_n,
\end{equation*}
and, at the node $n(k)$ of an observation, $\Sigma_k=\tilde H\ode{P}_n\tilde H^\top+\Robs$ and $P_n=\ode{P}_n-\ode{P}_n\tilde H^\top\Sigma_k^{-1}\tilde H\ode{P}_n$, and elsewhere $P_n=\ode{P}_n$.
Run with and without the observations, this recursion gives
\begin{equation*}
  \log|\operatorname{Cov}_\xi(Y\mid Z=0)| = \log|\operatorname{Cov}_\xi(Z,Y)|-\log|\operatorname{Cov}_\xi(Z)|
  = \sum_{n=1}^N\log|S_n|+\sum_{k=1}^K\log|\Sigma_k|-\sum_{n=1}^N\log|S^0_n|,
\end{equation*}
with $S_n^0$ the innovation covariances of the recursion without the observations.
With the force known, the same two recursions run on the state block alone, with $A_h$, $Q_h$, $C_n$ and $H$ in place of $\tilde A_h$, $\tilde Q_h$, $\tilde C_n$ and $\tilde H$, and the initial covariance of the state given the force, which is zero here (Sec.~\ref{sec:initialization}).
Marking their innovation covariances by a tilde,
\begin{equation*}
  I_\xi(U;Y\mid Z)=\frac12\bigg(\sum_{n=1}^N\log\frac{|S_n|}{|\tilde S_n|}+\sum_{k=1}^K\log\frac{|\Sigma_k|}{|\tilde\Sigma_k|}-\sum_{n=1}^N\log\frac{|S^0_n|}{|\tilde S^0_n|}\bigg),
  \qquad
  D_\xi=\frac12\sum_{k=1}^K\log\frac{|\Sigma_k|}{|\tilde\Sigma_k|}.
\end{equation*}
$I_\xi$ needs four recursions, with and without the observations and with the force unknown and known, and $D_\xi$ the two with the observations.
Each costs about as much as a filter pass, and its gradient in $\log\sigma_x^2$ follows by automatic differentiation.

\section{Diffusion sweeps}\label{app:state-diffusion}

\begin{figure}[ht]
  \centering
  \includegraphics[width=\textwidth]{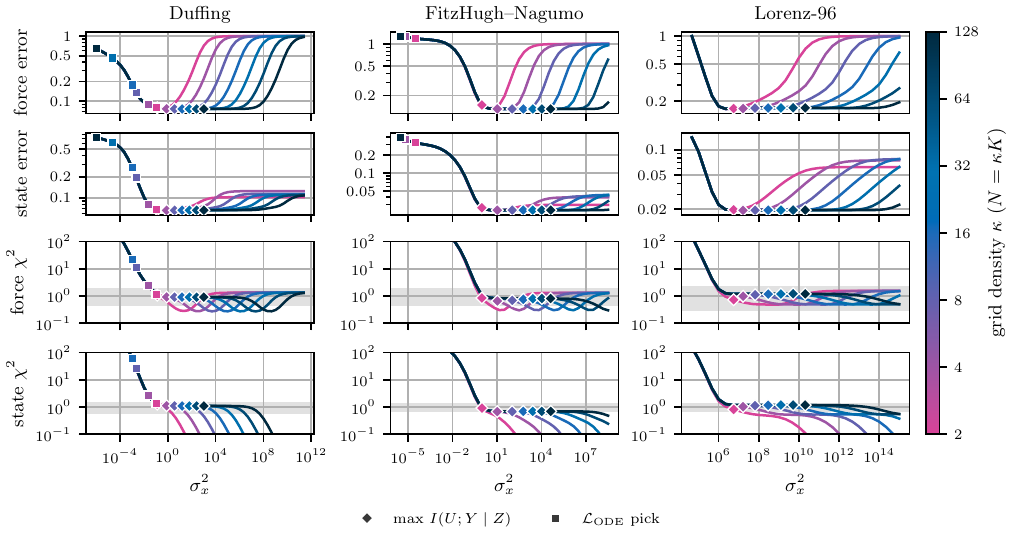}
  \caption{\textbf{Relative error and $\chi^2$ of force and state against the state diffusion.} Relative error of force and state (top rows) and their $\chi^2$ (bottom rows) against $\sigma_x^2$ on Duffing, FitzHugh--Nagumo and Lorenz-96 (columns) with $\varphi_u$ at the truth, one line per grid density $\kappa$, computed with the IEKS iterated to convergence. Only converged points are drawn. Diamonds: the $\sigma_x^2$ maximizing $I(U;Y\mid Z)$ on each grid (Fig.~\ref{fig:force-information}), placed on the curves by interpolation. Squares: the $\sigma_x^2$ maximizing $\lode$. Both in the grid's color.}
  \label{fig:sigma-x-sweep}
\end{figure}

\begin{figure}[ht]
  \centering
  \includegraphics[width=\textwidth]{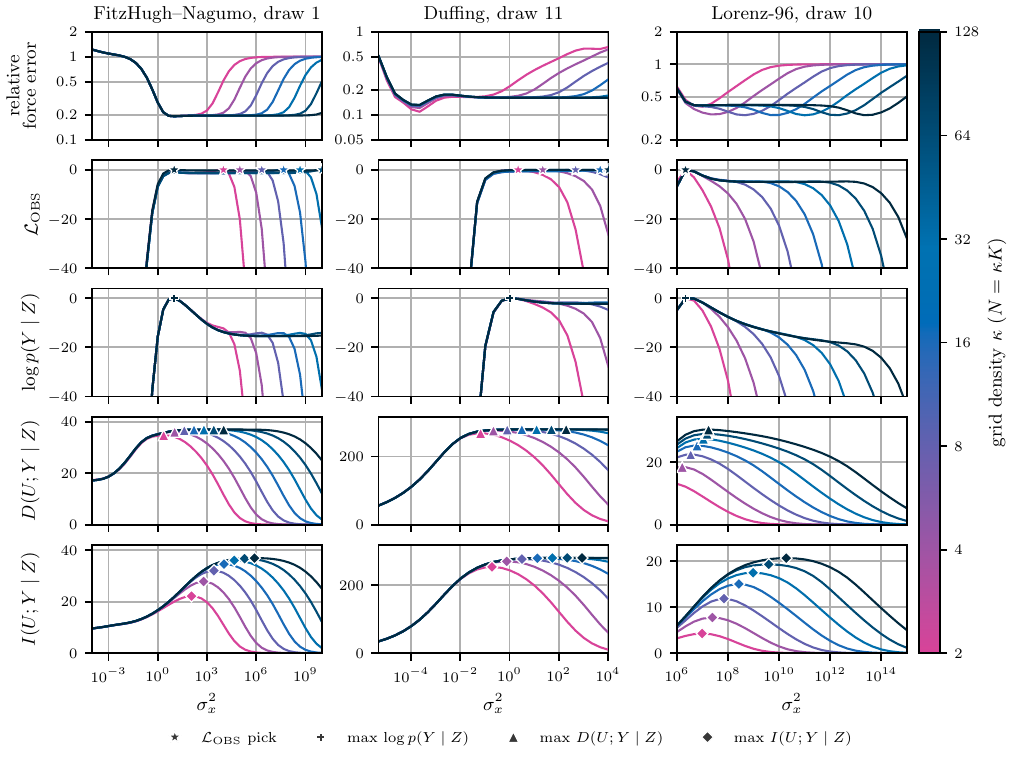}
  \caption{\textbf{Force error and the four criteria against the state diffusion on three problems that $\lobs$ misses.} Relative force error, $\lobs$ and $\log p(Y\mid Z)$ below their maxima, $D(U;Y\mid Z)$ and $I(U;Y\mid Z)$ (rows) against $\sigma_x^2$ on the first FitzHugh--Nagumo, the 11th Duffing and the 10th Lorenz-96 problem of Sec.~\ref{subsec:d123} (columns), with the IEKS and $\varphi_u$ at the truth, one line per grid density $\kappa$. Only converged points are drawn. Each criterion's row marks its own maximizer: stars for $\lobs$, plus signs for $\log p(Y\mid Z)$, triangles for $D(U;Y\mid Z)$ and diamonds for $I(U;Y\mid Z)$.}
  \label{fig:force-information}
\end{figure}

Figures~\ref{fig:failure-directions}, \ref{fig:force-information-duffing}, \ref{fig:sigma-x-sweep} and~\ref{fig:force-information} isolate the state diffusion: they hold the force hyperparameters $\varphi_u$ at the truth and compute the posterior with the IEKS.

\paragraph{Protocol.}\label{app:sweeps}
Figures~\ref{fig:force-information-duffing}, \ref{fig:sigma-x-sweep} and~\ref{fig:force-information} run the IEKS on a log-spaced grid of $\sigma_x^2$, separately on each grid density $\kappa$.
Figure~\ref{fig:failure-directions} runs $10$ IEKS iterations at each of its three diffusions on Duffing at $\kappa=32$.
A point counts as converged once the root-mean-square change of the scored state component over the last iteration is at most $10^{-3}$, and only converged points are drawn.
The sweep runs by continuation from the weak-enforcement end.
Each $\sigma_x^2$ starts from the converged trajectory of the next larger one, and from the EK1 smoothed mean at the largest value, after an unconverged point, or if the warm start does not converge.
An upward pass then restarts every unconverged point from its converged smaller neighbor.
Figure~\ref{fig:sigma-x-sweep} iterates in chunks of six IEKS iterations, at most six chunks per point, and scores at the interval midpoints.
Figures~\ref{fig:force-information-duffing} and~\ref{fig:force-information} run $30$ iterations per point and score at the observation times.
There the maximizers of $I_\xi$ and $D_\xi$ are refined by a parabola in $\log\sigma_x^2$ through the best grid point and its two neighbors and evaluated again, while the pick of $\lobs$ is its best grid point.
Figure~\ref{fig:failure-directions} uses the maximizer of $I_\xi$ on Duffing at $\kappa=32$.

\paragraph{Observation designs.}
The sweeps use earlier observation designs of the systems of Sec.~\ref{subsec:systems}, with the same dynamics and force priors.
On FitzHugh--Nagumo, $(v,w)(0)=(-1,-0.5)$ and $v$ is observed at $K=200$ times on $[0,200]$ with an SNR of $14$.
On Duffing, $(x,\dot x)(0)=(1,0)$ and the position $x$ is observed at $K=32$ times on $[0,62.5]$ with an SNR of about $20$.
On Lorenz-96, the sites $x_0,x_2,x_4,x_6$ are observed at $K=200$ times on $[0,10]$ with an SNR between $5.6$ and $9.0$.

\section{Implementation details}\label{app:implementation}

This appendix gives the parts of Algorithm~\ref{alg:calibration-guide} and of the experiments that the main text only names.
Filter and smoother are implemented in square-root form \citep{kraemer2024stable}, and all gradients are computed by automatic differentiation through them.

\subsection{Grid, initialization and search boxes}\label{sec:initialization}

On grid density $\kappa$, the $N=\kappa K$ steps of size $h=T/N$ have nodes $t_n=nh$, and Algorithm~\ref{alg:calibration-guide} doubles $\kappa$ up to $\kappa_{\max}=32$.
The observation times $t_k=kT/K$ are nodes, and $\kappa$ is even, so that the interval midpoints at which the posterior is scored are nodes as well.
The residual is linearized to first order (EK1), with $\Rode=0$.

\paragraph{Initial law} The force starts from the stationary distribution of its prior, $U(0)\sim\N(0,P^u_\infty)$, with $P^u_\infty$ the solution of the Lyapunov equation $F P^u_\infty + P^u_\infty F^\top + LL^\top = 0$ of the Matérn SDE with drift $F$ and dispersion $L$ (\citealp{sarkka2019applied} and Appendix~\ref{bsec:matern}).
The state starts from the known initial values $x(0)$, and $\dot x(0)$ for second-order systems.
Higher derivatives follow from differentiating the ODE at $t=0$ with Taylor-mode automatic differentiation \citep{kraemer2024stable}.
For an additive force these derivatives are affine in the force, $X(0)=m^x_0+J\,U(0)$, with $J$ the Jacobian of the Taylor coefficients of the state with respect to the force state, so that the joint initial distribution is exactly Gaussian,
\begin{equation*}
  \begin{bmatrix} X(0) \\ U(0) \end{bmatrix}
  \sim \N\!\left(\begin{bmatrix} m^x_0 \\ 0 \end{bmatrix},
  \begin{bmatrix} JP^u_\infty J^\top & JP^u_\infty \\ P^u_\infty J^\top & P^u_\infty \end{bmatrix}\right).
\end{equation*}
The state thus has no initial uncertainty of its own, and the ODE holds exactly at $t=0$ for every force.

\paragraph{Search boxes and start} The hyperparameters $\varphi=(\log\sigma_x^2,\log\sigma_u^2,\log\ell_u)$ are fitted in boxes.
The force variance $\sigma_u^2$, in units of the marginal variance of the true force, lies in $[e^{-20},e^{10}]$, and the length scale within a factor $e^{\pm2}$ of the truth.
The diffusion lies in $\sigma_x^2\in T^{-5}\,[10^{a},10^{b}]$, with $(a,b)=(3,13)$ on Duffing, $(6,20)$ on FitzHugh--Nagumo and Lorenz-96, and $(4,20)$ on the SIR model.
The start $\varphi^{(0)}$ lies at the fractions $0.8$, $0.7$ and $0.3$ of the boxes of $\log\sigma_x^2$, $\log\sigma_u^2$ and $\log\ell_u$, the same for every criterion: the diffusion on the weak-enforcement side, a large force amplitude $\sigma_u^2=e\,\sigma_u^{\star 2}$ and a short length scale $\ell_u\approx0.45\,\ell_u^\star$.

\subsection{Stability guard}\label{app:stability}

A filter pass counts as stable if the local error estimate of ODE filters \citep{bosch2021calibrated} stays below one at every step.
At step $n$, the residual is linearized with Jacobians $C_n$ and $B_n$ as in Appendix~\ref{app:update-equations}, and $\hat z_n$ is the linearized residual at the predicted mean $\pre{\mu}_n$.
It gives the local diffusion estimate and the local error on the components $\mathcal I$ of the ODE's state,
\begin{equation*}
  \hat s_n^2=\frac{1}{d}\,\hat z_n^\top\big(C_nQ_hC_n^\top+B_n\bar Q_hB_n^\top\big)^{-1}\hat z_n,
  \qquad
  \varepsilon_n=\bigg(\frac{1}{|\mathcal I|}\sum_{j\in\mathcal I}\frac{\hat s_n^2\,(Q_h)_{jj}}{\big(1+|(\pre{\mu_x})_{n,j}|\big)^2}\bigg)^{1/2},
\end{equation*}
with $d$ the dimension of the residual, and absolute and relative tolerance one.
The pass is stable if $\max_n\varepsilon_n\le1$.
The guard is checked on the EK1 pass at $\varphi^{(0)}$ that starts a grid (line~\ref{alg:kappa0}) and on the pass at the fitted $\varphi$ in every iteration of the IEKS.
After an unstable iteration the grid is discarded, and the next grid starts again from $\varphi^{(0)}$.

\subsection{Fitting at a frozen linearization}\label{app:fitting}

In each iteration of the IEKS, the residual is linearized at the trajectory $\xi$, and the hyperparameters are fitted at this frozen linearization by L-BFGS-B \citep{byrd1995limited} in their boxes, with at most $100$ iterations.
A non-finite value counts as $10^{10}$ with zero gradient.
The diffusion is fitted first, by the criterion $\mathcal J$, from its current value, and then $\varphi_u$ by $\lobs$.
For $\mathcal J=\lobs$ all three are fitted jointly.
All four criteria are maximized with their gradients.
The IEKS stops when $\lobs$ at the fitted $\varphi$ changes by less than $\tau_\ell=10^{-3}$ between iterations, or after $n_{\max}=5$ iterations.
A finer grid keeps the diffusion $\sigma_x^2$ of the previous grid's fit and its trajectory $\xi$, interpolated linearly in time, and restarts the force hyperparameters from $\varphi_u^{(0)}$.

\subsection{Grid acceptance}\label{app:acceptance}

Let $\mu^u_\kappa(t)$ and $s^u_\kappa(t)$ be the posterior mean and standard deviation of the force on grid $\kappa$.
The force estimate has settled if its change at the observation times is small against its posterior standard deviation at the interval midpoints $t_{k-1/2}$, in root mean square,
\begin{equation*}
  \bigg(\frac1K\sum_{k=1}^K\big(\mu^u_\kappa(t_k)-\mu^u_{\kappa/2}(t_k)\big)^2\bigg)^{1/2}
  < \tau_u\,\bigg(\frac1K\sum_{k=1}^K s^u_\kappa(t_{k-1/2})^2\bigg)^{1/2},
\end{equation*}
with $\tau_u=1$.
The fit reproduces the data through the ODE if $r_\kappa\le\tau_r=1.5$.
For $r_\kappa$, the observation times are split into windows of $W=16$ intervals, starting at the observation indices $s_w=W(w-1)$.
In window $w$, the ODE $\dot x=f(x,\mu^u_\kappa(t),t)$ is solved from the smoothed mean of the ODE's state at $t_{s_w}$ by the classical Runge--Kutta method with $32$ steps per observation interval, with the force mean interpolated linearly between the nodes.
With $\hat x(t_k)$ the solution at the observation times of its window, and $\sigma_j$ the noise standard deviation of channel $j$ of the $d_y$-dimensional observations,
\begin{equation*}
  r_\kappa=\bigg(\frac{1}{Kd_y}\sum_{k=1}^K\sum_{j=1}^{d_y}\Big(\frac{y_{k,j}-h_j\big(\hat x(t_k),\mu^u_\kappa(t_k)\big)}{\sigma_j}\Big)^2\bigg)^{1/2}.
\end{equation*}
A fit consistent with the data gives $r_\kappa\approx1$.
The check exposes a fit in which the numerical freedom of the state, not the force, carries the data: the ODE driven by the force mean then departs from them within a window even where the smoothed state follows them.
It does not always expose a force that is small against the autonomous dynamics over a window, such as the damped force of $\lobs$ on the 19th FitzHugh--Nagumo problem (Fig.~\ref{fig:failures-lobs}), which passes the check with $r_\kappa=1.39$.
If no grid is accepted, the last fit is reported, and the force is flagged as not identified.

\subsection{Baselines}\label{app:baselines}

\citet{tronarp2022fenrir} and \citet{wu2024data} as published evaluate $\log p(Y\mid Z)$ on their own one-pass EK1 filters at $\varphi$ rather than at a frozen IEKS linearization.
The former run a filter without the data and condition the backward Markov chain of its smoother on the data, the latter subtract $\lode$ of the filter without the data from $\lode+\lobs$ of the filter with the data.
At a shared linearization the two coincide, and in our experiments $\log p(Y\mid Z)$ is evaluated at the frozen IEKS linearization and maximized by L-BFGS-B with its gradient like the other criteria.
The classical latent force model \citep{hartikainen2012state} runs through Algorithm~\ref{alg:calibration-guide}, with the ODE filter replaced by the classical Runge--Kutta discretization of the ODE \citep{hairer1993solving}, treated as exact, and the force in the same Matérn state-space form.
It takes $\kappa\in\{1,2,4,8,16\}$ Runge--Kutta steps per observation interval, with the force evaluated at the stage times from the force state.
Without a state prior there is no diffusion to select, and a grid is accepted on the change of the force alone, against its posterior standard deviation at the observation times.
The ladder runs to the end and the accepted grid is read off afterwards, so its wall-clock time includes all five grids.
On Lorenz-96 an isotropic process noise on the state is fitted with $\varphi_u$, starting at the observation noise variance per observation interval, and kept from the coarser grid like the diffusion.
On odd grids the force at the midpoints is bridged between its neighboring nodes, and the state is not scored.

\subsection{Calibration interval of the \texorpdfstring{$\chi^2$}{chi-squared}}\label{app:chi2}

If the standardized errors $e_i/\sigma_g(t_i)$ at the $n$ scored midpoints are distributed as $\N(0,\rho)$ with a correlation matrix $\rho$, then $\chi^2_g$ is distributed as $\frac1n\sum_{j=1}^n\lambda_j\zeta_j^2$, with $\lambda_j$ the eigenvalues of $\rho$ and $\zeta_j\sim\N(0,1)$ independent.
Its distribution function follows from the formula of \citet{imhof1961computing},
\begin{equation*}
  P\Big(\sum_{j=1}^n\lambda_j\zeta_j^2\le x\Big)=\frac12-\frac1\pi\int_0^\infty\frac{\sin\theta(v)}{v\,\gamma(v)}\,\mathrm dv,\qquad
  \theta(v)=\frac12\sum_{j=1}^n\arctan(\lambda_jv)-\frac{xv}2,\qquad \gamma(v)=\prod_{j=1}^n\big(1+\lambda_j^2v^2\big)^{1/4},
\end{equation*}
a single integral, which we evaluate by quadrature and solve for the $2.5\,\%$ and $97.5\,\%$ quantiles.

The interval in Figs.~\ref{fig:comparison} and~\ref{fig:robustness} depends only on the design of a system, the number $n$ of scored midpoints and the true length scale $\ell_u^\star$, and is the same for every fit on it.
It takes the standardized force errors to be correlated like the Matérn-$5/2$ force prior,
\begin{equation*}
  \rho_{ij}=\Big(1+\frac{\sqrt5\,\tau_{ij}}{\ell_u^\star}+\frac{5\,\tau_{ij}^2}{3\,\ell_u^{\star 2}}\Big)\exp\Big(-\frac{\sqrt5\,\tau_{ij}}{\ell_u^\star}\Big),\qquad \tau_{ij}=|t_i-t_j|.
\end{equation*}
With midpoints a distance $T/n$ apart, the interval is set by $n$ and the ratio $\ell_u^\star/T$: the smoother the force relative to the window, the fewer independent errors and the wider the interval.


\end{document}

%% file: preamble.tex
\usepackage{amsmath}
\usepackage{amsthm}
\usepackage{amssymb} 
\usepackage{array}   
\usepackage{tikz}
\usepackage{algorithm}
\usepackage[noEnd=true,indLines=true]{algpseudocodex} 

\usetikzlibrary{arrows.meta, positioning, fit, backgrounds}
\definecolor{darkblue}{RGB}{26,58,91} 
\definecolor{tozero}{RGB}{0,114,178}
\definecolor{toinfty}{RGB}{213,94,0}
\newcommand{\tozero}[1]{\textcolor{tozero}{#1}}
\newcommand{\toinfty}[1]{\textcolor{toinfty}{#1}}

\theoremstyle{definition}

\theoremstyle{remark}

\DeclareRobustCommand{\couplingarrow}{\tikz[baseline=-0.55ex]{\draw[->, >={Latex[length=1.4mm]}, thick, densely dashed, darkblue] (0,0) -- (0.62,0);}}
\DeclareRobustCommand{\pickdiamond}{\tikz[baseline=-0.55ex]{\fill[darkblue] (0,-0.8mm) -- (0.8mm,0) -- (0,0.8mm) -- (-0.8mm,0) -- cycle;}}
\DeclareRobustCommand{\picksquare}{\tikz[baseline=-0.55ex]{\fill[darkblue] (-0.6mm,-0.6mm) rectangle (0.6mm,0.6mm);}}

\newcommand{\R}{\mathbb{R}}
\newcommand{\Rd}{\mathbb{R}^d}
\newcommand{\N}{\mathcal{N}}
\DeclareMathOperator*{\argmax}{arg\,max}

\newcommand{\lode}{\mathcal{L}_{\mathrm{ODE}}}
\newcommand{\lobs}{\mathcal{L}_{\mathrm{OBS}}}
\newcommand{\Rode}{R^{\mathrm{ODE}}}
\newcommand{\Robs}{R^{\mathrm{OBS}}}

\newcommand{\pre}[1]{#1^{-}}
\newcommand{\ode}[1]{#1^{\mathrm{+}}}

\hypersetup{hidelinks}
